\documentclass{article}

\PassOptionsToPackage{numbers, sort&compress}{natbib}
\usepackage[main, final]{neurips_2026}

\usepackage[utf8]{inputenc}
\usepackage[T1]{fontenc}
\usepackage{hyperref}
\usepackage{url}
\usepackage{booktabs}
\usepackage{amsfonts}
\usepackage{nicefrac}
\usepackage{microtype}
\usepackage{xcolor}
\definecolor{linkblue}{rgb}{0.21,0.49,0.74}
\hypersetup{colorlinks=true, allcolors=linkblue}
\usepackage{graphicx}
\usepackage{subcaption}

\usepackage{xspace}
\makeatletter
\DeclareRobustCommand\onedot{\futurelet\@let@token\@onedot}
\def\@onedot{\ifx\@let@token.\else.\null\fi\xspace}
\def\eg{\emph{e.g}\onedot}

\def\ie{\emph{i.e}\onedot}

\def\vs{\emph{vs}\onedot}

\makeatother

\usepackage{amsmath,amsfonts,bm}

\def\eqref#1{equation~\ref{#1}}

\def\1{\bm{1}}

\def\vc{{\bm{c}}}

\def\vg{{\bm{g}}}

\def\vo{{\bm{o}}}

\def\vs{{\bm{s}}}

\def\vw{{\bm{w}}}
\def\vx{{\bm{x}}}

\DeclareMathAlphabet{\mathsfit}{\encodingdefault}{\sfdefault}{m}{sl}
\SetMathAlphabet{\mathsfit}{bold}{\encodingdefault}{\sfdefault}{bx}{n}

\usepackage{times}
\usepackage{epsfig}
\usepackage{graphicx}
\usepackage{amsmath}
\usepackage{amssymb}
\usepackage{verbatim}
\usepackage{hhline}
\usepackage{multirow}
\usepackage{makecell}
\usepackage{booktabs}

\usepackage{listings}
\usepackage{xcolor}
\usepackage{caption}
\usepackage{tikzducks}
\usepackage{pgfplots}
\pgfplotsset{compat=1.18}
\usepackage{xcolor,colortbl}

\usepackage{mwe}
\usepackage{bbm}

\definecolor{grey}{rgb}{0.9, 0.9, 0.9}
\definecolor{electricviolet}{rgb}{0.56, 0.0, 1.0}
\newcommand{\dy}[1]{#1}
\definecolor{tealblue}{rgb}{0.0, 0.55, 0.55}
\newcommand{\dyt}[1]{#1}
\newcommand{\dyb}[1]{#1}
\newcommand{\upd}[1]{#1}
\newcommand{\dyn}[1]{#1}
\newcommand{\dyc}[1]{#1}
\newcommand{\ours}{LT-HOI}

\newcommand{\rcol}{\rowcolor{grey}}
\newcommand{\ccol}{\cellcolor{grey}}

\newcolumntype{L}[1]{>{\raggedright\let\newline\\\arraybackslash\hspace{0pt}}m{#1}}
\newcolumntype{C}[1]{>{\centering\let\newline\\\arraybackslash\hspace{0pt}}m{#1}}
\newcolumntype{R}[1]{>{\raggedleft\let\newline\\\arraybackslash\hspace{0pt}}m{#1}}

\providecommand{\and}{, }
\providecommand{\keywords}[1]{}

\usepackage{orcidlink}

\title{Long-Term Composition of \\ Human-Object Interactions}

\author{%
  Dayoung Gong$^{1}$ \qquad Minsu Cho$^{1,2}$ \qquad Gül Varol$^{3}$ \\ \\
  $^{1}$Pohang University of Science and Technology (POSTECH) \quad\quad 
  $^{2}$RLWRLD \\
  $^{3}$LIGM, École des Ponts, IP Paris, Univ Gustave Eiffel, CNRS \\
  {\small {\tt \url{https://gongda0e.github.io/LTHOI/}}}
}

\begin{document}

\maketitle

\begin{abstract}
Real-world human activities unfold as sequences of temporally dependent human-object interactions (HOIs), yet existing methods address either atomic HOIs in isolation or text-driven motion composition without object or scene context. To bridge this gap, we introduce the task of long-term action composition of HOIs, where a human navigates between and interacts with multiple objects within a 3D scene. The primary challenges lie in producing smooth transitions, scene-aware locomotion, and plausible hand-object interaction. To this end, we propose LT-HOI, a diffusion-based model that jointly conditions on past and future contexts—the past providing kinematic continuity across transitions, the future enabling anticipatory navigation toward upcoming interacting objects. We further integrate Diffusion Noise Optimization for collision avoidance and introduce a hand-object displacement loss to improve contact quality. On the ParaHome dataset, we establish the first benchmark for this task and demonstrate that LT-HOI composes long, temporally coherent HOIs.
\end{abstract}
\section{Introduction}
\label{sec:intro}

Human motion is inherently contextual, emerging from continuous interactions with the surrounding world. Real-world activities unfold not as isolated actions, but as sequences of temporally dependent human-object interactions (HOIs).
Each interaction builds upon the previous one—sharing object states, spatial configurations, and intentions—while transitioning smoothly to the next.
Capturing such temporal dependencies is essential for generating long-term and goal-oriented motion.

Despite recent progress in temporal action composition~\cite{athanasiou2022teach, zhang2023diffcollage, barquero2024seamless, zhaodartcontrol} and HOI generation~\cite{li2023object, li2024controllable, petrov2025tridi, xu2023interdiff, diller2024cg, jung2026decoupled, zhang2025bimart, wu2025human, li2026hoipage, petrov2026echo}, a significant gap remains. Temporal composition methods generate temporally aligned human-only motions from text without modeling objects or scenes, while HOI models focus on single short interactions.
This is partly due to the scarcity of long HOI sequences in existing datasets~\cite{li2023object, bhatnagar2022behave, taheri2020grab, huang2022intercap, jiang2023full}, leaving long-term HOI composition largely unexplored. The recently released ParaHome dataset~\cite{kim2025parahome} provides long-duration HOI sequences in 3D home environments, opening up the possibility of studying this problem.
\begin{figure}[t!]
    \centering
    \includegraphics[width=0.9\linewidth]{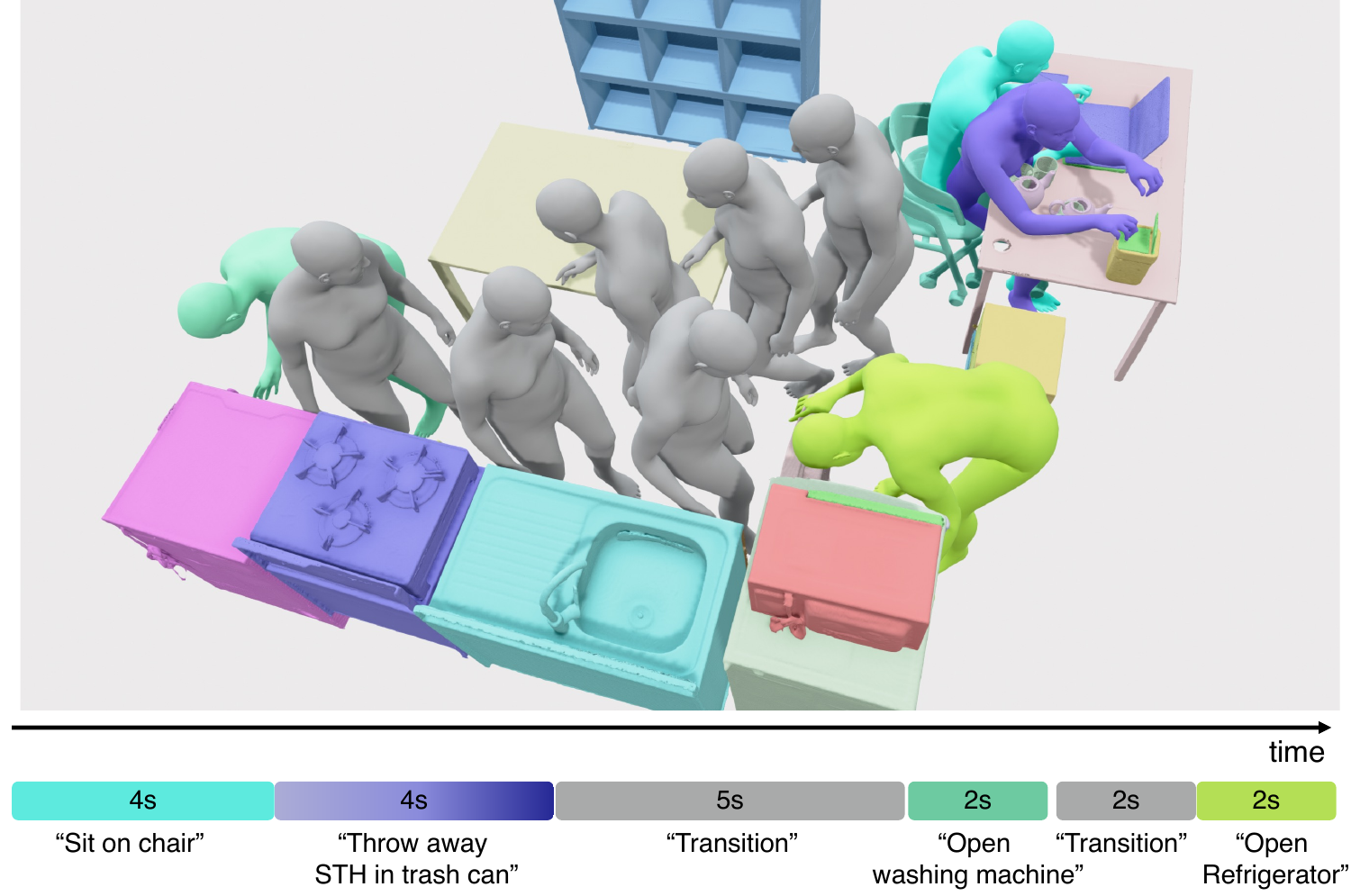}
    \vspace{-3mm}
    \caption{\textbf{Long-term action composition of HOIs} aims to generate temporally coherent HOI sequences within a 3D scene. The figure illustrates motions generated by our model. Different body colors denote different actions.
    }
    \label{fig:teaser}
    \vspace{-5mm}
\end{figure}

In this paper, we introduce the task of long-term action composition of HOIs: synthesizing motion sequences where a 3D human smoothly navigates between and interacts with multiple objects within a 3D scene (Fig.~\ref{fig:teaser}). We assume object motions are given as input and use them as spatio-temporal guidance for human motion synthesis. \dy{In practice, object motions can be obtained either from planning~\cite{yoshida2025generating, baik2026text} or from object pose estimation methods that recover object trajectories from human demonstrations or videos~\cite{wen2024foundationpose, wen2023bundlesdf}.} The primary challenges then lie in three aspects: (1) smooth transitions between consecutive interactions, (2) scene-aware locomotion toward upcoming interactions, and (3) plausible hand-object contact.

To address these challenges, we propose LT-HOI, a diffusion-based model that composes long-term HOI sequences.
At its core, \ours~jointly conditions on past and future contexts that play complementary roles: the past comprises previously generated human motion along with text, object motion, and object geometry, providing kinematic continuity for smooth transitions; the future comprises the upcoming text, object motion, and object geometry, providing spatial awareness for goal-directed locomotion.
In practice, we adopt an autoregressive-across-subsequences, non-autoregressive-within-subsequence generation scheme~\cite{athanasiou2022teach}. At each step, the model produces a motion segment covering the tail of the previous action, the current action, and the beginning of the next—naturally aligning with our past, current, and future context design. 
While future context guides the model to the right destination, navigating around scene obstacles requires explicit geometric reasoning. To this end, we further leverage Diffusion Noise Optimization (DNO)~\cite{karunratanakul2024optimizing} for collision-aware locomotion and introduce a hand-object displacement loss for plausible interaction.

Our contributions are as follows: \textbf{1)} We introduce the task of long-term action composition of HOIs and establish the first benchmark on the ParaHome dataset. \textbf{2)} We propose LT-HOI, a diffusion-based model that jointly conditions on past and future contexts to compose long HOI sequences, where past context ensures smooth transitions and future context enables goal-directed locomotion. \textbf{3)} We further integrate DNO to enable collision-aware navigation and introduce a hand-object displacement loss to improve contact quality during interactions. \textbf{4)} Comprehensive experiments validate the effectiveness of our approach, with ablations confirming the contribution of each component.

\section{Related Work}
\label{sec:realted_work}

\textbf{Temporal action composition for human motions.}
Given a series of text descriptions and their corresponding durations, temporal action composition aims to generate human motions that follow the instructions and exhibit coherent temporal structure.
TEACH~\cite{athanasiou2022teach} first introduces this task through a hierarchical formulation that generates motions non-autoregressively within segments and autoregressively across them, building upon a text-conditioned 3D motion synthesis model~\cite{petrovich2022temos}.
Subsequent works~\cite{shafir2023human, zhang2023diffcollage, barquero2024seamless} improve transition quality by refining overlapping segments—either via post-hoc inpainting~\cite{shafir2023human}, joint diffusion over temporal windows~\cite{zhang2023diffcollage}, or blended positional encodings~\cite{barquero2024seamless}. 
\dy{More recent works further explore streaming motion generation~\cite{xiao2025motionstreamer, ling2026prism}.}
However, these approaches focus on text-driven human-only motion composition and do not model interactions with objects or the surrounding environment.
In this paper, we address temporal composition of HOIs within a 3D scene.

\textbf{Human-object interactions.}
Modeling 3D human interactions with objects~\cite{li2023object, diller2024cg, xu2023interdiff, li2024controllable, petrov2025tridi, cao2025avatargo, song2024hoianimator} and scenes~\cite{lou2024harmonizing, huang2023diffusion, yi2024generating, Zhao:ICCV:2023} is a crucial task for embodied AI.
In this paper, we mainly focus on interacting with dynamic objects within a 3D scene. 
The goal of human-object interaction motion generation is to synthesize human motions that interact with objects.
Early methods such as OMOMO~\cite{li2023object} take a sequence of object states as input and generate full-body human motions that interact with the given object.
InterDiff~\cite{xu2023interdiff} introduces the task of anticipating future 3D human-object interactions, given past human and object motions. To achieve this, it employs an interaction diffusion model together with an interaction correction module.
HOI-Diff~\cite{peng2025hoi} proposes an affordance-guided diffusion model for HOI generation, while CG-HOI~\cite{diller2024cg} and TriDi~\cite{petrov2025tridi} model human, object, and contact interactions within a unified network.
Recently, CHOIS~\cite{li2024controllable} presented a diffusion-based framework that predicts both human and object motions conditioned on a text prompt, the initial human pose, the target object pose, object waypoints, and object geometries. \dy{DecHOI~\cite{jung2026decoupled} decouples trajectory generation from interaction modeling for HOI synthesis, and recent works further extend HOI generation to bimanual interactions with articulated objects~\cite{zhang2025bimart}, human-level instructions~\cite{wu2025human}, zero-shot generation with part affordance guidance~\cite{li2026hoipage}, egocentric modeling~\cite{petrov2026echo}, and large-scale mocap data~\cite{lu2025humoto}.} 
However, existing approaches primarily focus on single, short interactions, rather than multi-step, temporally coherent sequences involving multiple objects.
In this paper, we address long-term action composition of HOIs, aiming to generate temporally coherent human motions that sequentially interact with multiple objects within a 3D scene.

\section{Approach}

In this section, we present LT-HOI, a diffusion-based model for long-term action composition of HOIs. We formulate the generation as autoregressive across subsequences and non-autoregressive within each subsequence. At each step, the subsequence is generated conditioned on both past and future contexts---past context provides kinematic continuity for smooth transitions, and future context guides goal-directed locomotion toward the next interaction. We first describe the problem formulation (Sec.~\ref{sec:formulation}), data representation (Sec.~\ref{sec:data}), and the model architecture (Sec.~\ref{sec:model}), followed by training (Sec.~\ref{sec:training}) and inference (Sec.~\ref{sec:inference}) strategies.

\subsection{Problem Formulation}
\label{sec:formulation}

\dyc{Given the text descriptions, object geometries, and object motions of the interactions, together with a global scene map,} the goal is to synthesize human motions of $N$ consecutive interactions within a shared 3D scene. Specifically, the full motion sequence $\bm{X}$ consists of $N$ subsequences $\bm{X} = [\bm{X}_1, \cdots, \bm{X}_N]$, where each $\bm{X}_n = [\bm{x}_1, \cdots, \bm{x}_L] \in \mathbb{R}^{L \times D}$ represents a motion segment, $\bm{x}_l \in \mathbb{R}^D$ is the human pose at frame $l$, and $D$ is the pose dimension.
\dyc{Each subsequence $\bm{X}_n$ is generated by conditioning on three contexts, each consisting of text, object geometry, and object motion: the current context from its own interaction, and the past and future contexts from the previous and next interactions. The past context also includes the previously generated human motion.}

\subsection{Data Representation}
\label{sec:data}

\textbf{Human.}
We represent a human pose as $\bm{x} = [\bm{J}, \bm{r}, \bm{\theta}]$, where $\bm{J} \in \mathbb{R}^{N^\texttt{joint} \times 3}$ denotes the global joint positions, $\bm{r} \in \mathbb{R}^{6}$ is the 6D rotation~\cite{zhou2019continuity} of the global body orientation, and $\bm{\theta} \in \mathbb{R}^{(N^\texttt{joint}-1) \times 6}$ represents the 6D rotations of the SMPL-X~\cite{pavlakos2019expressive} pose parameters. $N^\texttt{joint}$ includes both body and hand joints, and all components are flattened and concatenated, yielding $D = 3N^\texttt{joint} + 6 + 6(N^\texttt{joint}-1)$.\\
\textbf{Object.}
The object pose $\bm{o} \in \mathbb{R}^{9}$ consists of its translation and 6D rotation. The object geometry $\bm{g} \in \mathbb{R}^{K}$ is encoded using Basis Point Sets (BPS)~\cite{prokudin2019efficient}, following~\cite{li2024controllable, li2023object}, where $K$ is the number of basis points. BPS provides a compact shape descriptor by measuring distances from a fixed set of basis points to the object surface.\\
\textbf{Scene.}
We represent the scene through three complementary maps projected onto the ground plane. First, a signed distance field $\bm{s}^{\texttt{sdf}} \in \mathbb{R}^{H \times W \times 1}$ encodes the walkable space around static obstacles, with values normalized from 0 (occupied) to 1 (free). Second, the initial positions of the current and future interacting objects are encoded as Gaussian heatmaps, $\bm{s}^{\texttt{curr}}, \bm{s}^{\texttt{future}} \in \mathbb{R}^{H \times W \times 1}$, respectively. The final scene representation $\bm{s} = [\bm{s}^{\texttt{sdf}}, \bm{s}^{\texttt{curr}}, \bm{s}^{\texttt{future}}] \in \mathbb{R}^{H \times W \times 3}$ captures both navigable geometry and the spatial layout of interactions.

\subsection{LT-HOI}
\label{sec:model}
\begin{figure*}[t!]
    \centering
    \includegraphics[width=\linewidth]{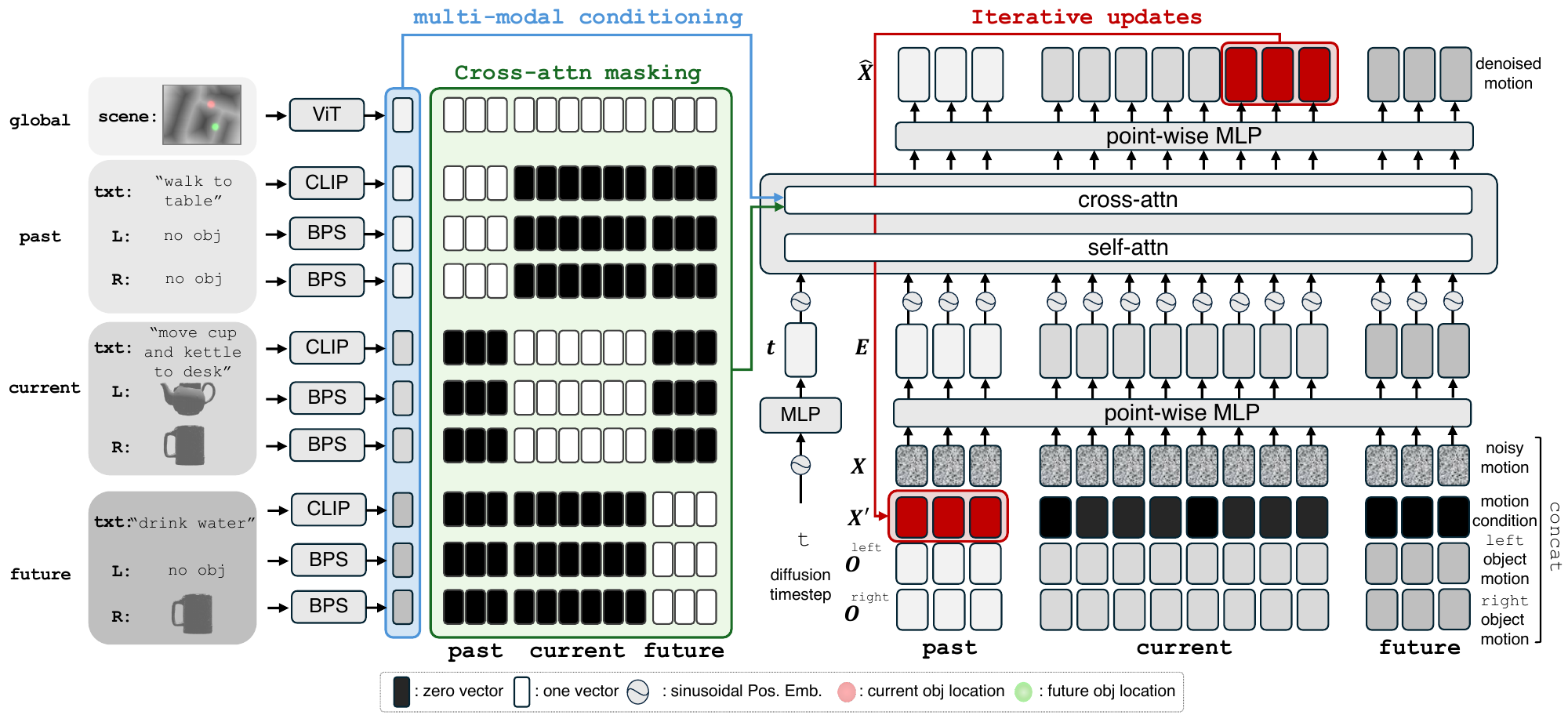}
    \caption{\textbf{Overall framework of \ours.} \ours~is a diffusion-based generative model for long-term HOI composition. The model generates HOIs autoregressively across subsequences and non-autoregressively within each subsequence, conditioning on past and future context while iteratively feeding previous predictions into subsequent generation.
    }
    \label{fig:pipeline}
    \vspace{-3mm}
\end{figure*}
The overall pipeline is illustrated in Fig.~\ref{fig:pipeline}. At each autoregressive step, the model generates a single subsequence $\bm{X}_n$ of total length $L^{\texttt{all}}$, composed of three temporally distinct segments: the tail of the previous motion $\bm{X}_n^{\texttt{past}}$ (length $L^{\texttt{past}}$), the current target motion $\bm{X}_n^{\texttt{curr}}$ (length $L^{\texttt{curr}}$), and the beginning of the next motion $\bm{X}_n^{\texttt{future}}$ (length $L^{\texttt{future}}$). The full subsequence is $\bm{X}_n = [\bm{X}_n^{\texttt{past}}, \bm{X}_n^{\texttt{curr}}, \bm{X}_n^{\texttt{future}}] \in \mathbb{R}^{L^{\texttt{all}} \times D}$, where $L^{\texttt{all}} = L^{\texttt{past}} + L^{\texttt{curr}} + L^{\texttt{future}}$.

\textbf{Input embedding.}
Since each temporal segment involves a distinct action with potentially different objects, we encode the text description and object geometry per segment $p \in \{\texttt{past}, \texttt{curr}, \texttt{future}\}$. Text descriptions are encoded with CLIP~\cite{radford2021learning} and projected through MLPs to obtain text embeddings $\bm{w}^p \in \mathbb{R}^{C}$. BPS representations of the left and right object geometries, $\bm{g}^{\texttt{left},p}, \bm{g}^{\texttt{right},p} \in \mathbb{R}^{K}$, are similarly projected to $\tilde{\bm{g}}^{\texttt{left},p}, \tilde{\bm{g}}^{\texttt{right},p} \in \mathbb{R}^{C}$, with learnable \texttt{[left]} and \texttt{[right]} indicator tokens added via summation. The scene map $\bm{s}$ is patchified and processed by a Vision Transformer (ViT)~\cite{dosovitskiy2021vit}, with the resulting \texttt{[CLS]} token projected to obtain the scene feature $\tilde{\bm{s}} \in \mathbb{R}^{C}$. To facilitate smooth transitions, we construct a motion condition $\bm{X}'_n \in \mathbb{R}^{L^{\texttt{all}} \times D}$, where the past segment is filled with the clean previous motion $\bm{X}_n^{\texttt{past}}$ and the remaining positions are zero-padded. The diffusion timestep $t$ is encoded via a sinusoidal positional embedding followed by an MLP, yielding $\bm{t} \in \mathbb{R}^{C}$.

\textbf{Multimodal conditioning.}
We integrate conditions based on their temporal granularity. Frame-wise conditions---the per-frame object motions $\bm{O}^{\{\texttt{left}, \texttt{right}\}}$ and the past motion condition $\bm{X}'_n$---are concatenated channel-wise with the noisy motion $\bm{X}_n$ and projected through an MLP to produce the motion embedding:
\begin{equation}
    \bm{E} = \texttt{MLP}([\bm{X}_n, \bm{X'}_n, \bm{O}^{\texttt{left}}, \bm{O}^{\texttt{right}}]).
\end{equation}
This channel-wise integration provides dense spatial correspondence between the human and objects at every frame. In contrast, subsequence-wise conditions---text descriptions $\bm{w}^p$ and object geometries $\tilde{\bm{g}}^p$ for each segment $p$---and the global scene feature $\tilde{\bm{s}}$ are incorporated via cross-attention. These are assembled into \dyc{condition tokens} $\bm{M}$.

\textbf{Denoising transformer.}
The motion embedding $\bm{E}$, together with the timestep embedding $\bm{t}$, is fed into a denoising transformer $f$ consisting of $N^{\texttt{dec}}$ decoder layers. Each layer applies self-attention over the motion sequence followed by cross-attention to $\bm{M}$. During cross-attention, a subsequence-aware mask restricts each motion frame to attend only to condition tokens of its corresponding segment, while the scene token is globally accessible across all segments. The transformer output $\bm{F} \in \mathbb{R}^{L^{\texttt{all}} \times C}$ is:
\begin{equation}
    \bm{F} = f(\bm{E}, \bm{t};\, \bm{M}).
\end{equation}
The final predicted motion $\hat{\bm{X}}_n \in \mathbb{R}^{L^{\texttt{all}} \times D}$ is obtained by a point-wise MLP:
\begin{equation}
    \hat{\bm{X}}_n = \texttt{MLP}(\bm{F}).
\end{equation}
For brevity, we denote the entire denoising network—comprising input embedding (Eq.~1), denoising transformer (Eq.~2), and output projection (Eq.~3)—as $f_\theta$, so that $\hat{\bm{X}}_n = f_\theta(\bm{X}_n, t; \bm{M})$.

\subsection{Training}
\label{sec:training}

The model is trained via the standard diffusion denoising objective. During the forward process, noise is added to the clean motion $\bm{X}^0$ at a randomly sampled timestep $t$:
\begin{equation}
    \bm{X}^t = \sqrt{\gamma(t)}\bm{X}^{0} + \sqrt{1-\gamma(t)}\epsilon, \quad \epsilon \sim \mathcal{N}(0, \bm{I}),
\end{equation}
where $\gamma(t)$ is a pre-defined variance schedule. 
The denoising network $f_\theta$ is trained to recover $\bm{X}^0$ from $\bm{X}^t$ using an $\ell_1$ reconstruction loss:
\begin{equation}
    \mathcal{L}^{\text{recon}} = \|f_\theta(\bm{X}^t, t; \bm{M}) - \bm{X}^0\|_1.
\end{equation}

\textbf{Hand-object displacement loss.}
To improve the quality of hand-object interactions, we introduce an auxiliary loss that supervises the spatial relationship between hand joints and object surfaces.
For each interacting object, we first obtain a set of $P$ sparse representative points $\bm{Q}\in\mathbb{R}^{P\times 3}$ by applying Farthest Point Sampling (FPS)~\cite{qi2017pointnetplusplus} to the object mesh vertices, which are then transformed per frame using the ground-truth object pose.
Given $N^\texttt{hand}$ hand joint positions $\bm{J}^\texttt{hand}\in\mathbb{R}^{N^\texttt{hand}\times 3}$, we define the BPS feature function $\phi(\bm{J}^\texttt{hand}, \bm{Q})\in\mathbb{R}^{N^\texttt{hand}\times 3}$ as the displacement vector from each hand joint to its nearest point in $\bm{Q}$.
The displacement loss is then defined as the $\ell_1$ distance between the BPS features computed from the predicted hand joints $\hat{\bm{J}}^\texttt{hand}$ and those from the ground-truth hand joints $\bm{J}^\texttt{hand}$:
\begin{align}
    \mathcal{L}^\texttt{disp} = \| \phi(\hat{\bm{J}}^\texttt{hand}, \bm{Q}) - \phi(\bm{J}^\texttt{hand}, \bm{Q}) \|_1.
    \label{eq:contact_loss}
\end{align}
This loss is computed independently for the left and right hands with their respective interacting objects, and is applied only at frames where an object interaction is present.
By supervising the relative spatial relationship between hands and objects rather than absolute hand positions, the loss encourages more plausible hand-object contacts.

The final training objective is $\mathcal{L} = \lambda^\texttt{recon}\mathcal{L}^\texttt{recon} + \lambda^\texttt{disp}\mathcal{L}^\texttt{disp}$, where $\lambda^\texttt{recon}$ and $\lambda^\texttt{disp}$ control the weights of the respective losses.

\subsection{Inference}
\label{sec:inference}

The model generates the full long-term sequence by autoregressively synthesizing $N$ subsequences during inference. The process begins with no past context ($L^{\texttt{past}} = 0$) for the first subsequence. After generating $\bm{X}_n$, the last $L^{\texttt{past}}$ frames of $\bm{X}_n^{\texttt{curr}}$ serve as the past context for the next step, forming $\bm{X}_{n+1}^{\texttt{past}}$. The canonical frame for the first subsequence is set using the ground-truth yaw orientation; for subsequent steps, it is derived from the predicted motion.

Within each subsequence, motion is synthesized using the reverse diffusion process. Starting from pure Gaussian noise $\hat{\bm{X}}^S$, the denoising network $f_\theta$ iteratively refines the sequence, $\hat{\bm{X}}^S \rightarrow \hat{\bm{X}}^{S-\Delta} \rightarrow \cdots \rightarrow \hat{\bm{X}}^0$,
following the deterministic update rule of DDIM~\cite{song2020denoising}, where $\Delta$ denotes the step size that skips intermediate diffusion steps\dy{ to accelerate sampling}.

\textbf{Obstacle-avoidance navigation using DNO.}
To enable obstacle-avoidance navigation without compromising the learned motion prior, we adopt Diffusion Noise Optimization (DNO)~\cite{karunratanakul2024optimizing}, a test-time optimization framework that steers a pre-trained diffusion model toward a target criterion by updating its initial latent noise rather than the model parameters or the motion itself. We instantiate this criterion as a scene-collision loss defined on the pelvis trajectory following DNO~\cite{karunratanakul2024optimizing}. 
Specifically, at each optimization iteration, we first run a full reverse-diffusion pass starting from the initial noise $\hat{\bm{X}}^S$ to obtain the predicted clean motion $\hat{\bm{X}}^0$. We then extract the pelvis trajectory from $\hat{\bm{X}}^0$, project it onto the 2D ground plane, and query the scene SDF to compute the collision loss:
\begin{equation}
    \mathcal{L}^\texttt{col} = \frac{1}{L^\texttt{curr} + L^\texttt{future}} \sum_{l=L^\texttt{past}+1}^{L^\texttt{all}} c_l \;+\; \max_{l}\, c_l,
    \label{eq:obs_loss}
\end{equation}
where $c_l = \max\!\left(0,\; \delta - \bm{s}^\texttt{sdf}(\hat{\bm{J}}^\texttt{pel}_l)\right)$ is the per-frame collision penalty, $\hat{\bm{J}}^\texttt{pel}_l$ is the projected pelvis position at frame $l$, and $\delta$ is a safety margin. The first term penalizes the average collision over the current and future segments, and the second penalizes the worst-case collision. The gradient of $\mathcal{L}^\texttt{col}$ is then backpropagated through the entire reverse-diffusion process to update $\hat{\bm{X}}^S$.
Since the update modifies only the latent noise while keeping the diffusion model fixed, the resulting locomotion stays close to the learned motion prior while reducing scene collisions.

\section{Experiments}

We first describe the dataset (Sec.~\ref{sec:dataset}), evaluation metrics (Sec.~\ref{sec:metrics}), and implementation details (Sec.~\ref{sec:impl}). We then compare \ours~with state-of-the-art methods (Sec.~\ref{sec:sota}), present analyses on the generation strategy, individual components, and the role of past and future contexts (Sec.~\ref{sec:analysis}), and provide qualitative results (Sec.~\ref{sec:qualitative}).

\subsection{ParaHome Dataset}
\label{sec:dataset}

We use the ParaHome dataset~\cite{kim2025parahome} to evaluate long-term composition of HOI motion generation. The dataset contains 207 scenes collected from 38 participants, where each participant interacts with 22 objects, resulting in a total of 486 minutes of human motions performing daily activities in real home environments. Each scene includes, on average, 26 sequences of HOI actions and 39 sequences of transition actions between HOI interactions. Transition actions refer to the in-between motions connecting consecutive HOI actions---including locomotion where the human moves across the 3D scene to reach the next interaction---which are crucial for evaluating temporal smoothness. We additionally annotate these motions as `Transition'.\\
\textbf{Splits.}
Since our primary goal is to evaluate the temporal composition of motions, and each scene in the dataset contains consecutive motion sequences, we split the dataset by scene to prevent data leakage across splits. The training, validation, and test sets consist of 165, 21, and 21 scenes, respectively, maintaining an approximate 8:1:1 ratio. In total, these splits contain 5,970, 1,005, and 989 \dyb{subsequences}, respectively. \dy{All sequences are processed at 30 FPS. Training subsequences average 113.5 frames (3.8\,s), and \dyb{each test scene is generated as a single continuous sequence by composing} approximately 47 consecutive subsequences on average (4,556 frames, about 2.5 minutes of continuous HOIs). Notably, 13.7\% of all subsequences are locomotion without any object interaction, where the model must navigate toward the next interaction using only the text, scene, and future context.}\\
\textbf{Interacting objects.}
Since the dataset does not provide annotations for interacting objects, we automatically detect them from the data based on hand-object proximity. Specifically, we obtain human vertices using the SMPL-X model~\cite{pavlakos2019expressive} and compute distances between hand vertices and object vertices. If the minimum distance falls below a predefined threshold of 0.02\,m, the object is regarded as interacting. Detection is performed independently for the left and right hands.
Since ParaHome represents articulated objects with multiple parts---for example, a gas stove is composed of a main body and two knobs, a laptop consists of top and bottom parts, and a washing machine consists of a main body and a door---we adopt part-level object representations, where each part has its own translation and rotation. Non-articulated objects are represented as a single part. Each object part is encoded with object geometry $\bm{g}$ and object motions $\bm{O}$.

\subsection{Evaluation Metrics}
\label{sec:metrics}

We evaluate the performance of our model from three perspectives: subsequence, transition, and sequence, following previous work~\cite{barquero2024seamless, li2024controllable, hwang2025scenemimotioninbetweeningmodeling}.\\
\textbf{Subsequence.}
For subsequence evaluation, we assess human motion, goal reaching, and contact quality. Following prior single-HOI generation work~\cite{li2024controllable}, we use MPJPE, foot height, foot sliding (FS), R@3, and FID for human motion, and contact precision, recall, F1, and hand penetration for contact quality. We further use translation error ($T_\text{goal}$) and orientation error ($O_\text{goal}$) to evaluate how accurately the human reaches the target object at the moment of interaction.\\
\textbf{Transition.}
To evaluate the motion smoothness when transitioning between consecutive $\text{HOI}$ actions, we define and evaluate the transition motions. Specifically, the transition motion $\bm{X}^{\texttt{tr}} \in \mathbb{R}^{L^{\texttt{tr}} \times D}$ is defined as the motion segment of length $L^{\texttt{tr}}$ that bridges the termination of subsequence $\bm{X}_n$ and the start of $\bm{X}_{n+1}$.
As evaluation metrics for transition motions, we adopt FID, PeakJerk (PJ), and Area Under the Jerk (AUJ)~\cite{barquero2024seamless}.\\
\textbf{Sequence.}
To evaluate scene-aware locomotion, we further report the scene collision ratio (Coll$_\%$) and the maximum scene penetration depth (P$_\mathrm{max}$) following ~\cite{hwang2025scenemimotioninbetweeningmodeling}, which respectively measure the fraction of frames in which the body interpenetrates the 3D scene and the worst-case penetration depth into scene geometry.
\noindent
Detailed explanations of metrics are provided in Sec.~\ref{sec_supp:eval_metrics}.

\subsection{Implementation Details}
\label{sec:impl}

We use $N^\texttt{joint} = 52$ joints in total, including 22 body joints and 30 hand joints. Global joint positions are extracted using the SMPL-X joint regressor~\cite{pavlakos2019expressive}. All human, object, and scene data are canonicalized with respect to the yaw orientation of the human pose at the first frame of the current subsequence. Joint positions and object translations are then normalized to the range $[-1, 1]$.
The denoising transformer consists of $N^{\texttt{dec}} = 8$ decoder layers with 4 attention heads and a channel dimension of $C = 256$. We train the model with a batch size of 32 and an initial learning rate of $2.5 \times 10^{-4}$, using a linear warm-up over the first 2,000 iterations followed by cosine decay. The total number of training iterations is 200,000. During inference, we use \dy{75 DDIM steps and 150 DNO optimization steps. We set} the boundary margin $\delta$ to 0.15\,m.  The transition length $L^{\texttt{tr}}$ is set to 30 frames. We set $\lambda^\texttt{recon} = 1$ and $\lambda^\texttt{disp} = 0.3$.

\subsection{Comparison with the State of the Art}
\label{sec:sota}
\begin{table*}[t]
\caption{\textbf{Comparison with the State of the Art} }
\label{tab:sota}
\centering
\setlength{\tabcolsep}{3pt}
\resizebox{\textwidth}{!}{
\begin{tabular}{lccccccccccccccccc}
\toprule
\multirow{3}{*}{Method} &
\multicolumn{12}{c}{\textbf{Subsequence}} &
\multicolumn{3}{c}{\textbf{Transition}} &
\multicolumn{2}{c}{\textbf{Sequence}} \\
\cmidrule(lr){2-13}\cmidrule(lr){14-16}\cmidrule(lr){17-18}
&
\multicolumn{7}{c}{\textbf{Motion}} &
\multicolumn{5}{c}{\textbf{Contact}} &
\multicolumn{1}{c}{\textbf{Motion}} &
\multicolumn{2}{c}{\textbf{Jerk}} &
\multicolumn{2}{c}{\textbf{Scene}} \\
\cmidrule(lr){2-8}\cmidrule(lr){9-13}\cmidrule(lr){14-14}\cmidrule(lr){15-16}\cmidrule(lr){17-18}
&
MPJPE$\downarrow$ & H$_\mathrm{feet}\rightarrow$ & FS$\downarrow$ &
R@$3\uparrow$ & FID$\downarrow$ &
T$_\mathrm{goal}\downarrow$ & O$_\mathrm{goal}\downarrow$ &
C$_\mathrm{prec}\uparrow$ & C$_\mathrm{rec}\uparrow$ & C$_\mathrm{F1}\uparrow$ & C$_\%\rightarrow$ & P$_\mathrm{hand}\downarrow$ &
FID$\downarrow$ & PJ$\downarrow$ & AUJ$\downarrow$ &
Coll$_\%\downarrow$ & P$_\mathrm{max}\downarrow$ \\
\midrule
GT & - & -3.30 & 0.29 & - & - & - & - & - & - & - & 0.49 & 1.7e-6 & - & 0.03 & 2.22 & 0.35 & 0.04 \\
\midrule
MDM & \dyt{28.76} & \dyt{-6.39} & \dyt{0.33} & \dyt{0.43} & \dyt{0.70} & \dyt{54.14} & \upd{1.04} & \dyt{0.34} & \dyt{0.18} & \dyt{0.21} & \dyt{0.18} & \dyt{6.2e-5} & \dyt{1.59} & \dyt{0.16} & \textbf{\dyt{2.05}} & \dyt{0.67} & \dyt{0.07} \\
CHOIS & \dyt{28.47} & \dyt{-6.90} & \dyt{0.34} & \dyt{0.44} & \dyt{1.22} & \dyt{54.81} & \upd{1.07} & \dyt{0.33} & \dyt{0.18} & \dyt{0.20} & \dyt{0.17} & \dyt{3.4e-5} & \dyt{1.21} & \dyt{0.17} & \dyt{2.09} & \dyt{0.60} & \dyt{0.06} \\
\dy{DecHOI} & \upd{27.26} & \textbf{\upd{-3.43}} & \upd{0.39} & \dyt{0.44} & \dyt{0.53} & \upd{50.33} & \upd{0.98} & \upd{0.40} & \dyt{0.23} & \dyt{0.26} & \upd{0.24} & \upd{8.7e-5} & \dyt{2.44} & \upd{0.21} & \upd{2.23} & \upd{0.58} & \upd{0.06} \\
InterDiff & 23.17 & -5.17 & \textbf{0.28} & 0.36 & 1.09 & 39.85 & 0.65 & 0.38 & 0.20 & 0.23 & 0.19 & \textbf{2.5e-6} & 4.60 & 0.80 & 3.44 & 0.51 & 0.06 \\
FlowMDM & 24.46 & \dyt{-4.44} & 1.03 & 0.46 & 0.64 & 53.25 & 0.71 & 0.41 & 0.24 & 0.27 & 0.24 & 3.2e-6 & 6.16 & \textbf{0.12} & \dyt{2.12} & 0.53 & 0.06 \\
\rcol \ours~(ours) & \textbf{\dy{19.16}} & \dy{-4.28} & \dy{0.33} & \textbf{\dy{0.54}} & \textbf{\dy{0.45}} & \textbf{\dy{24.41}} & \textbf{\dy{0.48}} & \textbf{\dy{0.49}} & \textbf{\dy{0.36}} & \textbf{\dy{0.38}} & \textbf{\dy{0.35}} & \dy{8.2e-5} & \textbf{\dy{1.12}} & \dy{0.13} & \dy{2.16} & \textbf{\dy{0.32}} & \textbf{\dy{0.05}} \\
\bottomrule
\end{tabular}
}
\end{table*}

\dy{We compare \ours~with MDM~\cite{tevethuman}, CHOIS~\cite{li2024controllable}, DecHOI~\cite{jung2026decoupled}, InterDiff~\cite{xu2023interdiff}, and FlowMDM~\cite{barquero2024seamless}, which target text-conditioned motion generation, HOI synthesis (CHOIS and DecHOI), HOI prediction, and temporal action composition, respectively.
For a fair comparison, all baselines are retrained from scratch on ParaHome with the same input modalities and training recipe as \ours: MDM and FlowMDM additionally take object motion, object geometry, and scene features; CHOIS and DecHOI take scene features; and InterDiff takes text and scene features.
Other design choices, including the network architecture, follow the original works: MDM, CHOIS, and DecHOI generate each subsequence conditioned on the last frame of the previous one, following their original setting that takes the initial human pose; InterDiff predicts a subsequence from past observations; and FlowMDM jointly generates the full sequence.
Implementation details are provided in Sec.~\ref{sec_supp:implementation_details}.}

\dy{Table~\ref{tab:sota} presents the results. \ours~outperforms the baselines on most metrics, and the gap reflects how each baseline handles temporal context.
Observing only a single past frame, MDM, CHOIS, and DecHOI cannot anticipate where the next interaction occurs, resulting in large goal-reaching errors (over 50\,cm vs.\ 24.41\,cm for \ours). DecHOI improves over MDM and CHOIS in MPJPE and contact quality, but its MPJPE remains higher than those of InterDiff, FlowMDM, and \ours, suggesting that strong atomic HOI synthesis alone does not address long-term composition.
InterDiff conditions on past motion but, without future context, has higher goal-reaching error and lower contact quality than \ours~(T$_\mathrm{goal}$ 39.85 vs.\ 24.41, C$_\mathrm{F1}$ 0.23 vs.\ 0.38).
FlowMDM avoids subsequence boundaries by generating the entire sequence jointly and thus achieves low transition jerk, yet its transition FID is several times higher than \ours~(6.16 vs.\ 1.12), indicating that low jerk alone does not yield realistic transitions.
By leveraging both past and future contexts, \ours~achieves smooth transitions, accurate goal reaching, and better contact quality.}

\subsection{Analysis}
\label{sec:analysis}
\begin{table*}[t]
\caption{\textbf{Effect of the generation strategy.} Generating each subsequence in parallel and composing subsequences autoregressively outperforms both variants on most metrics.}
\label{tab:variants}
\centering
\setlength{\tabcolsep}{3pt}
\resizebox{\textwidth}{!}{
\begin{tabular}{lccccccccccccccccc}
\toprule
\multirow{3}{*}{Method} &
\multicolumn{12}{c}{\textbf{Subsequence}} &
\multicolumn{3}{c}{\textbf{Transition}} &
\multicolumn{2}{c}{\textbf{Sequence}} \\
\cmidrule(lr){2-13}\cmidrule(lr){14-16}\cmidrule(lr){17-18}
&
\multicolumn{7}{c}{\textbf{Motion}} &
\multicolumn{5}{c}{\textbf{Contact}} &
\multicolumn{1}{c}{\textbf{Motion}} &
\multicolumn{2}{c}{\textbf{Jerk}} &
\multicolumn{2}{c}{\textbf{Scene}} \\
\cmidrule(lr){2-8}\cmidrule(lr){9-13}\cmidrule(lr){14-14}\cmidrule(lr){15-16}\cmidrule(lr){17-18}
&
MPJPE$\downarrow$ & H$_\mathrm{feet}\rightarrow$ & FS$\downarrow$ &
R@$3\uparrow$ & FID$\downarrow$ &
T$_\mathrm{goal}\downarrow$ & O$_\mathrm{goal}\downarrow$ &
C$_\mathrm{prec}\uparrow$ & C$_\mathrm{rec}\uparrow$ & C$_\mathrm{F1}\uparrow$ & C$_\%\rightarrow$ & P$_\mathrm{hand}\downarrow$ &
FID$\downarrow$ & PJ$\downarrow$ & AUJ$\downarrow$ &
Coll$_\%\downarrow$ & P$_\mathrm{max}\downarrow$ \\
\midrule
Joint & 22.61 & \textbf{-4.24} & 0.48 & 0.41 & 3.35 & 37.19 & 0.56 & 0.35 & 0.23 & 0.23 & 0.23 & \textbf{4.0e-6} & 12.84 & 0.82 & 3.35 & 0.57 & 0.06 \\
Autoregressive & 35.23 & 0.37 & 0.42 & 0.15 & 55.91 & 121.49 & 1.30 & 0.03 & 0.01 & 0.01 & 0.01 & 4.4e-6 & 25.30 & 0.64 & 2.95 & 0.68 & 0.06 \\
\rcol \ours & \textbf{\dy{19.16}} & \dy{-4.28} & \textbf{\dy{0.33}} & \textbf{\dy{0.54}} & \textbf{\dy{0.45}} & \textbf{\dy{24.41}} & \textbf{\dy{0.48}} & \textbf{\dy{0.49}} & \textbf{\dy{0.36}} & \textbf{\dy{0.38}} & \textbf{\dy{0.35}} & \dy{8.2e-5} & \textbf{\dy{1.12}} & \textbf{\dy{0.13}} & \textbf{\dy{2.16}} & \textbf{\dy{0.32}} & \textbf{\dy{0.05}} \\
\bottomrule
\end{tabular}
}
\end{table*}
\vspace{-2mm}

\begin{table*}[t!]
\caption{\textbf{Ablation study.} DNO reduces scene collisions and improves goal reaching, while $\mathcal{L}^{\texttt{disp}}$ improves contact quality.}
\label{tab:ablation}
\centering
\setlength{\tabcolsep}{3pt}
\resizebox{\textwidth}{!}{
\begin{tabular}{lcccccccccccccccccccc}
\toprule
\multirow{3}{*}{Method} &
\multirow{3}{*}{DNO} &
\multirow{3}{*}{$\mathcal{L}^\mathrm{disp}$} &
\multicolumn{12}{c}{\textbf{Subsequence}} &
\multicolumn{3}{c}{\textbf{Transition}} &
\multicolumn{2}{c}{\textbf{Sequence}} \\
\cmidrule(lr){4-15}\cmidrule(lr){16-18}\cmidrule(lr){19-20}
& & &
\multicolumn{7}{c}{\textbf{Motion}} &
\multicolumn{5}{c}{\textbf{Contact}} &
\multicolumn{1}{c}{\textbf{Motion}} &
\multicolumn{2}{c}{\textbf{Jerk}} &
\multicolumn{2}{c}{\textbf{Scene}} \\
\cmidrule(lr){4-10}\cmidrule(lr){11-15}\cmidrule(lr){16-16}\cmidrule(lr){17-18}\cmidrule(lr){19-20}
& & &
MPJPE$\downarrow$ & H$_\mathrm{feet}\rightarrow$ & FS$\downarrow$ &
R@$3\uparrow$ & FID$\downarrow$ &
T$_\mathrm{goal}\downarrow$ & O$_\mathrm{goal}\downarrow$ &
C$_\mathrm{prec}\uparrow$ & C$_\mathrm{rec}\uparrow$ & C$_\mathrm{F1}\uparrow$ & C$_\%\rightarrow$ & P$_\mathrm{hand}\downarrow$ &
FID$\downarrow$ & PJ$\downarrow$ & AUJ$\downarrow$ &
Coll$_\%\downarrow$ & P$_\mathrm{max}\downarrow$ \\
\midrule
\ours & \ccol $\checkmark$ & \ccol $\checkmark$ & \ccol \textbf{\dy{19.16}} & \ccol \textbf{\dy{-4.28}} & \ccol \dy{0.33} & \ccol \textbf{\dy{0.54}} & \ccol \dy{0.45} & \ccol \textbf{\dy{24.41}} & \ccol \textbf{\dy{0.48}} & \ccol \textbf{\dy{0.49}} & \ccol \textbf{\dy{0.36}} & \ccol \textbf{\dy{0.38}} & \ccol \textbf{\dy{0.35}} & \ccol \dy{8.2e-5} & \ccol \dy{1.12} & \ccol \dy{0.13} & \ccol \dy{2.16} & \ccol \textbf{\dy{0.32}} & \ccol \textbf{\dy{0.05}} \\
 & - & $\checkmark$ & 19.85 & \dy{-4.47} & 0.35 & \textbf{0.54} & \textbf{\dy{0.41}} & 25.70 & \upd{0.52} & 0.46 & \dy{0.35} & 0.36 & 0.34 & 8.6e-5 & 0.81 & 0.13 & \textbf{\dy{2.15}} & 0.45 & \textbf{\dy{0.05}} \\
 & - & - & 20.65 & -5.87 & \textbf{0.30} & 0.52 & 0.54 & 26.58 & \upd{0.56} & 0.43 & 0.28 & 0.30 & 0.27 & \textbf{6.5e-5} & \textbf{0.43} & \textbf{0.09} & \textbf{\dy{2.15}} & 0.48 & \textbf{\dy{0.05}} \\
\bottomrule
\end{tabular}
}
\end{table*}

\begin{figure*}[t!]
    \centering
    \includegraphics[width=\linewidth]{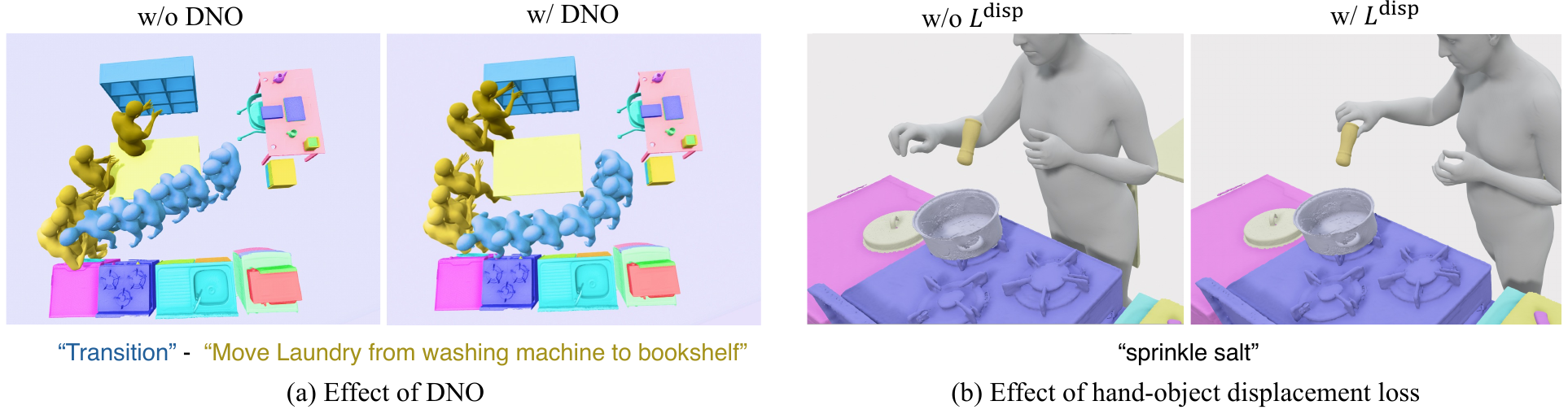}
    \caption{\textbf{Ablation study on DNO and $\mathcal{L}^{\texttt{disp}}$} }
    \label{fig:ablation}
    \vspace{-3mm}
\end{figure*}

\begin{figure*}[t]
\centering
\includegraphics[width=\textwidth]{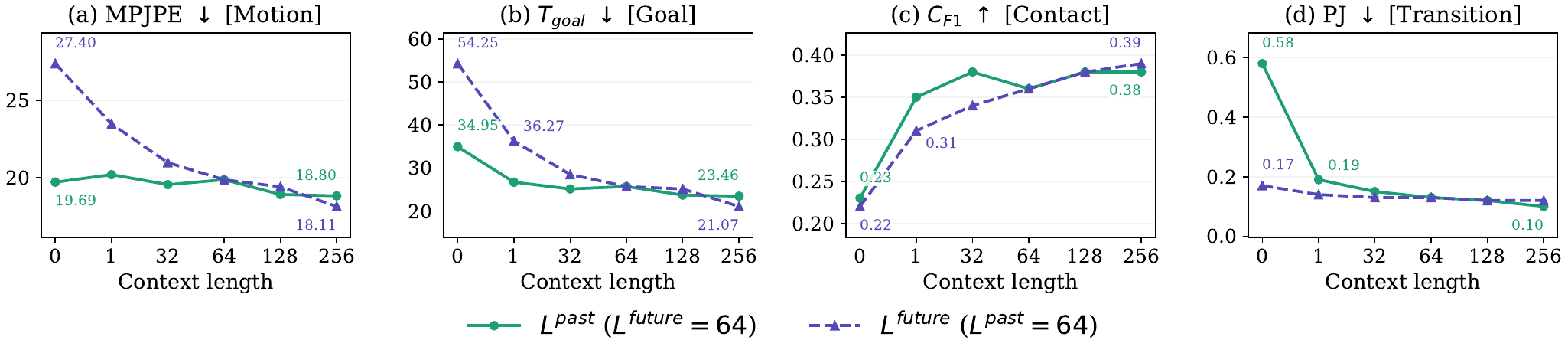}
\caption{\textbf{Effect of past and future context length.} Past context drives transition smoothness (d), future context drives motion accuracy (a) and goal reaching (b), while contact F1 (c) benefits from both contexts.}
\label{fig:context_ablation}
\end{figure*}

\textbf{Effect of the generation strategy.}
To validate our generation scheme, which produces HOI motions in parallel within each subsequence and autoregressively across subsequences, we compare against two variants: a \textit{joint} variant that generates the entire sequence in parallel, and an \textit{autoregressive} (AR) variant that generates motions sequentially in fixed-length chunks. For the AR variant, we use a chunk size of 32 frames. Table~\ref{tab:variants} presents the results.\\
\ours~achieves the best results on most metrics among the three schemes. The AR variant suffers from long-horizon error accumulation, \dy{showing high errors} on goal-reaching metrics ($T_{\text{goal}}$ 121.49, $O_{\text{goal}}$ 1.30), which in turn limits contact quality ($C_{\text{F1}}$ 0.01). The joint variant avoids this issue by modeling the entire sequence at once, but produces less precise interactions ($C_{\text{F1}}$ 0.23, FID 3.35), likely because jointly modeling all subsequences makes it harder to capture fine-grained per-segment details. \ours~combines the strengths of both: parallel generation within each subsequence for local coherence, and autoregressive composition across subsequences for long-horizon consistency.

\textbf{Ablation study.}
Table~\ref{tab:ablation} progressively ablates DNO and the hand-object displacement loss $\mathcal{L}^{\texttt{disp}}$.
Without DNO, the collision ratio (Coll$_\%$) increases from 0.32 to 0.45, and the goal-reaching error increases (T$_\mathrm{goal}$ 24.41 $\rightarrow$ 25.70).
As illustrated in Fig.~\ref{fig:ablation}(a), DNO enables collision-avoiding navigation. \dy{Further analyses of DNO, including per-body-part collisions, the objective design, and runtime, are provided in Sec.~\ref{sec_supp:dno}.}
Further removing $\mathcal{L}^{\texttt{disp}}$ degrades contact quality. 
Figure~\ref{fig:ablation}(b) shows that without $\mathcal{L}^{\texttt{disp}}$ the generated hands hover near the object without making proper contact, whereas applying it guides the hands toward the object surface for more plausible interactions.

\textbf{Effect of past context.}
We analyze the effect of past context length $L^{\texttt{past}}$ on temporal composition quality by varying it from 0 to 256 while fixing $L^{\texttt{future}} = 64$, as shown in Fig.~\ref{fig:context_ablation}. Without any past context ($L^{\texttt{past}} = 0$), the peak jerk reaches 0.58, indicating abrupt transitions between consecutive subsequences. This is expected, as the model has no information about how the previous motion ended, forcing it to generate the beginning of each segment independently. Introducing even a single frame of past context ($L^{\texttt{past}} = 1$) reduces PJ from 0.58 to 0.19, indicating that even minimal past information substantially improves transition smoothness.
As $L^{\texttt{past}}$ further increases, PJ continues to decrease, reaching 0.10 at $L^{\texttt{past}} = 256$, suggesting that longer past context provides richer kinematic history for the model to maintain temporal continuity. Beyond transition quality, longer past context also improves motion quality (MPJPE from 19.69 to 18.80), contact quality (F1 from 0.23 to 0.38), and goal-reaching accuracy ($\text{T}_{\text{goal}}$ from 34.95 to 23.46). This indicates that richer temporal history helps the model better coordinate not only smooth transitions, but also hand-object interactions and spatial navigation. \dy{Past context thus improves the coherence of compositions.}

\textbf{Effect of future context.}
We examine the effect of future context length $L^{\texttt{future}}$ by varying it from 0 to 256 while fixing $L^{\texttt{past}} = 64$, as shown in Fig.~\ref{fig:context_ablation}. The results reveal that future context plays a different role from past context. Without future context ($L^{\texttt{future}} = 0$), the largest degradation appears in goal-reaching accuracy: $\text{T}_{\text{goal}}$ rises to 54.25, indicating that the model struggles to navigate toward the next interacting object. Contact quality also drops (F1 of 0.22), as the generated motion lacks awareness of where the upcoming interaction will occur and thus cannot position the body appropriately.
Providing a single frame of future context ($L^{\texttt{future}} = 1$) already reduces $\text{T}_{\text{goal}}$ from 54.25 to 36.27 and improves contact F1 from 0.22 to 0.31, indicating that even a minimal spatial cue about the next interaction target is informative for navigation. As $L^{\texttt{future}}$ increases further, \dyb{MPJPE, contact F1, and goal-reaching accuracy continue to improve}: MPJPE decreases from 27.40 to 18.11, contact F1 increases from 0.22 to 0.39, and $\text{T}_{\text{goal}}$ drops to 21.07 at $L^{\texttt{future}} = 256$.\\
These results suggest that past and future contexts serve complementary roles. Past context mainly provides kinematic continuity by stabilizing transitions at segment boundaries, while future context provides spatial awareness of the upcoming interaction, guiding navigation toward the correct location for the next interaction. Combining both leads to improved overall performance, as each addresses a different aspect of long-term motion coherence.

\begin{table}[t]
\caption{\textbf{Robustness to object motion perturbations.} Metrics remain close to the clean setting ($\sigma=0$) up to $\sigma=5$\,cm, and contact recall decreases at $\sigma=10$\,cm.}
\label{tab:robustness}
\centering
\footnotesize
\setlength{\tabcolsep}{2.5pt}
\begin{tabular}{cccccccccccccc}
\toprule
$\sigma$ (cm) & $p$ (\%) & T$_\mathrm{obj}$ (cm) & MPJPE$\downarrow$ & FS$\downarrow$ & R@$3\uparrow$ & FID$\downarrow$ & T$_\mathrm{goal}\downarrow$ & C$_\mathrm{prec}\uparrow$ & C$_\mathrm{rec}\uparrow$ & C$_\mathrm{F1}\uparrow$ & FID$_\mathrm{tr}\downarrow$ & PJ$\downarrow$ & Coll$_\%\downarrow$ \\
\midrule
0 & 0 & 0.00 & 19.85 & 0.35 & 0.54 & 0.41 & 25.70 & 0.46 & 0.35 & 0.36 & 0.81 & 0.13 & 0.45 \\
\midrule
\multicolumn{14}{l}{\textit{(A) Temporal coverage}} \\
5 & 10 & 0.55 & 19.80 & 0.34 & 0.50 & 0.44 & 25.70 & 0.46 & 0.36 & 0.37 & 0.72 & 0.12 & 0.46 \\
5 & 25 & 1.33 & 19.93 & 0.34 & 0.51 & 0.43 & 25.84 & 0.46 & 0.34 & 0.36 & 0.77 & 0.13 & 0.46 \\
5 & 50 & 2.68 & 19.98 & 0.34 & 0.52 & 0.43 & 26.09 & 0.46 & 0.34 & 0.36 & 0.66 & 0.13 & 0.47 \\
5 & 75 & 4.19 & 19.91 & 0.33 & 0.53 & 0.42 & 26.42 & 0.46 & 0.33 & 0.36 & 1.17 & 0.13 & 0.46 \\
5 & 100 & 5.33 & 19.80 & 0.34 & 0.53 & 0.43 & 25.23 & 0.45 & 0.34 & 0.35 & 1.00 & 0.13 & 0.47 \\
\midrule
\multicolumn{14}{l}{\textit{(B) Noise magnitude}} \\
0.5 & 100 & 0.53 & 19.71 & 0.34 & 0.53 & 0.41 & 25.64 & 0.46 & 0.35 & 0.36 & 0.82 & 0.13 & 0.46 \\
1 & 100 & 1.07 & 19.74 & 0.35 & 0.52 & 0.39 & 25.20 & 0.46 & 0.35 & 0.37 & 0.87 & 0.13 & 0.47 \\
2 & 100 & 2.13 & 19.62 & 0.33 & 0.53 & 0.39 & 25.35 & 0.46 & 0.35 & 0.36 & 1.18 & 0.13 & 0.47 \\
5 & 100 & 5.33 & 19.80 & 0.34 & 0.53 & 0.43 & 25.23 & 0.45 & 0.34 & 0.35 & 1.00 & 0.13 & 0.47 \\
10 & 100 & 10.67 & 20.28 & 0.34 & 0.51 & 0.48 & 26.59 & 0.46 & 0.29 & 0.33 & 0.98 & 0.13 & 0.48 \\
\bottomrule
\end{tabular}
\end{table}

\dyn{\textbf{Robustness to object motion perturbations.}
Since \ours~takes object motions as input, we evaluate its robustness to imperfect object trajectories. At inference time, we perturb object translations with temporally smooth random 3D offsets (low-pass filtered noise) of magnitude $\sigma$ over a portion $p$ of the trajectory, with DNO disabled; T$_\mathrm{obj}$ denotes the resulting mean object translation error. As shown in Table~\ref{tab:robustness}, the metrics remain comparable to the clean setting even when the entire trajectory is perturbed with $\sigma=5$\,cm (contact F1 0.35 vs.\ 0.36). Under $\sigma=10$\,cm, contact recall decreases (0.35~$\rightarrow$~0.29) while precision is unchanged. Full results are provided in Sec.~\ref{sec_supp:robustness}.}

\begin{figure}[t!]
    \centering
    \includegraphics[width=\linewidth]{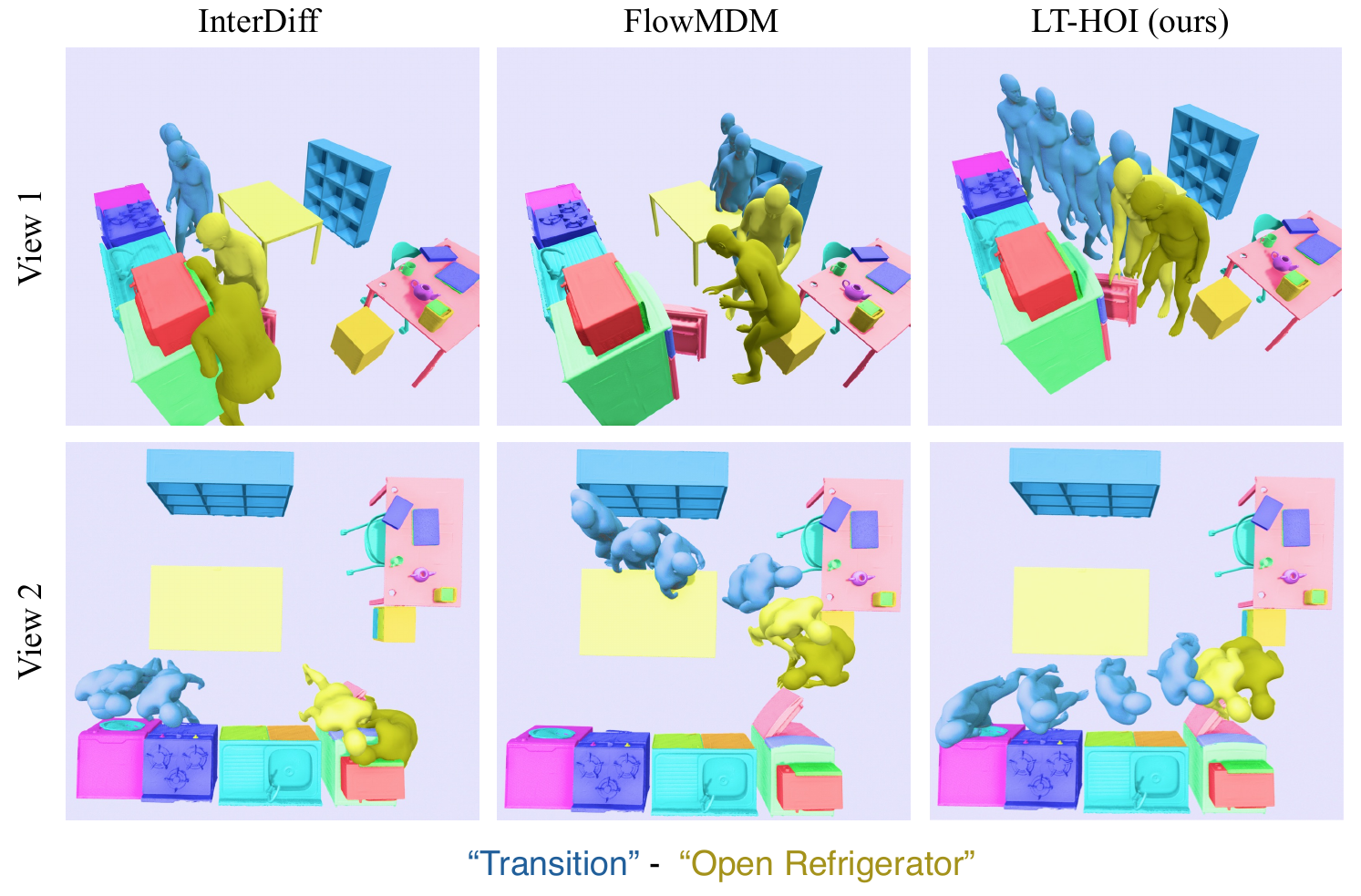}
    \vspace{-3mm}
    \caption{\textbf{Qualitative results} \dy{on a sequence of `Transition' followed by `Open Refrigerator', shown from two viewpoints with 20 frames uniformly sampled across the sequence. }}
    \label{fig:sota}
    \vspace{-5mm}
\end{figure}
\subsection{Qualitative Results}
\label{sec:qualitative}
Figure~\ref{fig:sota} presents qualitative comparisons between \ours~and the baselines. The example sequence consists of a `Transition' motion followed by an `Open Refrigerator' motion, with 20 frames uniformly sampled across the sequence. 
InterDiff reaches the next interaction location but skips the walking transition in between, resulting in an abrupt jump rather than continuous motion. FlowMDM produces reasonable locomotion but penetrates the desk. In contrast, \ours~generates smooth transitions between interaction locations while avoiding collisions with surrounding objects, producing plausible interaction motions throughout the sequence.

\dyn{\noindent\textbf{Perceptual study.}
We further conducted a perceptual study with 25 participants on 20 randomly sampled test sequences, where participants compared \ours~against InterDiff or FlowMDM in terms of semantic correctness, visual realism, and physical plausibility. \ours~is preferred over InterDiff in 95.2--96.0\% and over FlowMDM in 70.4--80.0\% of the comparisons across all three criteria, and the preference over FlowMDM rises to 89.5\% when participants perceive a clear difference (Sec.~\ref{sec_supp:user_study}). This suggests that the quantitative improvements of \ours~translate into motions that humans perceive as more faithful to the task, more natural, and more physically plausible.}

\vspace{-2mm}
\section{Conclusion}
\label{sec:conclusion}

We have introduced the task of long-term action composition of HOIs, aiming to generate human motions that coherently connect multiple object manipulations within a 3D scene.
We have proposed a diffusion-based generative model dubbed \ours, which synthesizes temporally aligned motions conditioned on past and future contexts—each comprising text, object motion, and object geometry. Past context enables smooth transitions between interactions, while future context guides goal-directed, scene-aware locomotion toward upcoming objects.
Experiments on the ParaHome dataset demonstrate that our framework synthesizes long, temporally coherent human motions. We hope this work provides a useful step toward goal-oriented, long-horizon motion generation for embodied AI.

\section*{Acknowledgements}
This work was supported by the Institute of Information \& Communications Technology Planning \& Evaluation (IITP) grants funded by the Korean government (MSIT) (RS-2024-00457882: National AI Research Lab Project (50\%), RS-2022-II220264: Comprehensive Video Understanding and Generation with Knowledge-based Deep Logic Neural Network (45\%), RS-2019-II191906: AI Graduate School Program at POSTECH (5\%)). 
This work also benefited from Hi! PARIS and State funding managed by the ANR under the France 2030 program (reference: ANR-23-IACL-0005), as well as the ANR project CorVis (ANR-21-CE23-0003-01). 
We thank Jeonghwan Kim, Jisoo Kim, and Hanbyul Joo for their assistance in reproducing the baseline implementations on the ParaHome dataset.

\bibliographystyle{abbrvnat}
\bibliography{main}

\appendix
\clearpage
\section*{Appendix}

\suppressfloats[t]
\appendix
\setcounter{table}{0}
\renewcommand{\thetable}{S\arabic{table}}
\setcounter{figure}{0}
\renewcommand{\thefigure}{S\arabic{figure}}

\vspace{0.5em}

In this supplementary material, we provide evaluation metrics in Sec.~\ref{sec_supp:eval_metrics}, \dy{details of the baselines, their past and future conditioned variants,} \dyb{and their short-term HOI synthesis results} in Sec.~\ref{sec_supp:implementation_details}, additional experimental results in Sec.~\ref{sec_supp:exp} \dy{(including additional ablation studies, scene representation, robustness to object motion perturbations, DNO analysis, and results on the HIMO dataset),} \dyb{a perceptual study in Sec.~\ref{sec_supp:user_study},} \dy{and discussions in Sec.~\ref{sec_supp:discussions}}.

\section{Evaluation Metrics}
\label{sec_supp:eval_metrics}
In this section, we provide detailed explanations of evaluation metrics used in our experiments (Sec.~\ref{sec:metrics}).
Following the previous work on atomic HOI~\cite{li2024controllable} and temporal action composition of human motions~\cite{barquero2024seamless}, we evaluate the generated motions in three parts---subsequence, transition, and sequence---using the following metrics. The symbols $\uparrow$ and $\downarrow$ indicate that higher or lower values correspond to better performance, respectively.
The symbol $\rightarrow$ indicates that the closer the value is to GT, the better.\\
\textbf{Mean per joint position error (MPJPE) $\downarrow$:} MPJPE measures the average Euclidean distance between the predicted and GT 3D joint positions, reported in centimeters. \\
\textbf{Foot height $\rightarrow$:} Foot height measures the foot height from the ground, reported in centimeters.\\
\textbf{Foot sliding $\downarrow$:} The foot sliding is computed as the weighted accumulated foot translation on the ground plane (xy-plane), following prior work~\cite{he2022nemf, li2024controllable}, reported in centimeters. \\
\textbf{R@3 $\uparrow$:} R@3 evaluates motion-to-text retrieval accuracy within the top-3 candidates, following Guo et al.~\cite{guo2022generating}.  It measures the capability of a motion representation to be semantically aligned with its corresponding natural language description. A higher $\text{R@3}$ score indicates better motion-text correspondence and representation quality.
For evaluation, we adhere to the standard protocol by training separate motion and text encoders on the training split of the dataset.
This training procedure follows the official implementation of Guo et al.~\cite{guo2022generating}. \\
\textbf{Fr\'echet Inception Distance (FID) $\downarrow$:} FID measures the distributional discrepancy between the predicted motions and the ground-truth motions in the learned feature space. We leverage the same motion encoder \dyn{trained on the ParaHome training split} for the $\mathbf{R@3}$ evaluation, ensuring consistency in the feature space used for distribution assessment. \\
\noindent\textbf{Goal reaching translation error ($T_\text{goal}$) $\downarrow$}: $T_\text{goal}$ measures how accurately the human reaches the target object at the moment of interaction. It is computed as the Euclidean distance between the human pelvis position and the target object position at the first frame of contact, reported in centimeters.\\
\noindent\textbf{Goal reaching orientation error ($O_\text{goal}$) $\downarrow$}: $O_\text{goal}$ measures the angular alignment between the human body and the target object at the moment of interaction. It is computed as the angular difference (in radians) between the body's forward direction and the direction toward the target object position, evaluated at the first frame of contact.\\
\textbf{Contact:} We evaluate hand–object interaction quality using contact precision \textbf{C$_\text{prec} \uparrow$}, recall \textbf{C$_\text{rec} \uparrow$}, F1 score \textbf{C$_\text{F1} \uparrow$}, and contact percentage \textbf{C$_\%\rightarrow$}~\cite{li2023object, li2024controllable}.\\
Contact is defined per frame by computing the minimum distance between each wrist joint (left and right) and all object vertices; if the minimum distance is below a fixed threshold, the frame is considered to be in contact. The same procedure is applied to both ground-truth and predicted motions, and we compute precision, recall, and F1 based on the resulting binary contact labels. Contact percentage measures the proportion of frames in which contact occurs, providing an overall estimate of how frequently the hand interacts with the object throughout the sequence.\\
\noindent
\textbf{P$_\text{hand}\downarrow$:} Hand penetration distance measures the average depth of hand–object interpenetration using a signed distance field (SDF). Specifically, the score is computed as the mean absolute value of all negative signed distances. Note that the penetration score is computed for the objects that interact with the hands.
\\
\textbf{Peak jerk (PJ)~$\downarrow$:} PJ measures abrupt motion changes during transitions. The metric is defined as the maximum jerk value observed across all joints during the transition motion, where jerk is the third derivative of joint position with respect to time~\cite{barquero2024seamless}.\\
\textbf{Area under jerk (AUJ) $\downarrow$:} AUJ evaluates the overall motion smoothness of the transition motion. It is computed as the difference between the maximum jerk across all joints and the average dataset jerk value, and this difference is further averaged over all transition motions~\cite{barquero2024seamless}.\\
\textbf{Scene collision ratio (Coll$_\%$)~$\downarrow$:} Coll$_\%$ quantifies how often the generated human body interpenetrates the static 3D scene. We voxelize each scene into a 3D signed distance field (SDF) and, for every frame, query the SDF at the body vertices obtained from the SMPL-X forward pass. To exclude trivial floor contact, only vertices above a fixed floor height of 10\,cm are considered. A frame is counted as colliding when at least one queried vertex has a signed distance below $-3$\,cm, i.e., a penetration depth larger than 3\,cm. Coll$_\%$ is then computed as the fraction of colliding frames over the total sequence length and averaged across all test sequences~\cite{hwang2025scenemimotioninbetweeningmodeling}.\\
\textbf{Scene penetration max (P$_\mathrm{max}$)~$\downarrow$:} P$_\mathrm{max}$ measures the worst-case penetration depth between the body and the scene. Using the same 3D SDF query as Coll$_\%$, the per-frame penetration depth is defined as the maximum of $\max(-d, 0)$ over all body vertices, where $d$ denotes the signed distance to the scene. P$_\mathrm{max}$ takes the maximum of these per-frame values across the entire sequence and is averaged across test sequences, reported in meters. While Coll$_\%$ captures how frequently collisions occur, P$_\mathrm{max}$ reflects how severe the most extreme collision is~\cite{hwang2025scenemimotioninbetweeningmodeling}.

\section{\dy{Comparison with Baselines}}
\label{sec_supp:implementation_details}

We compare \ours~against \dy{five} diffusion baselines: MDM~\cite{tevethuman}, CHOIS~\cite{li2024controllable}, \dy{DecHOI~\cite{jung2026decoupled},} InterDiff~\cite{xu2023interdiff}, and FlowMDM~\cite{barquero2024seamless}.
\dy{\subsection{Implementation Details}}
\label{sec_supp:baseline_impl}
For a fair comparison, all baselines share the motion representation, diffusion schedule (cosine, $1{,}000$ steps, $x_0$-prediction), CLIP ViT-B/32 text encoder, BPS encoder, ViT scene-BEV encoder, and transformer hyperparameters ($d_\text{model}{=}256$, $4$ heads, $8$ layers, FFN size $1{,}024$) of \ours.
The shared conditioning inputs are a per-segment text embedding $\vw_s\!\in\!\mathbb{R}^{512}$, per-segment left/right object BPS $\vg^{L}_s,\vg^{R}_s\!\in\!\mathbb{R}^{1024\times 3}$, a per-segment scene BEV image $\vs_s\!\in\!\mathbb{R}^{3\times 224\times 224}$, and per-frame object motion $\vo^{L}_t,\vo^{R}_t\!\in\!\mathbb{R}^{9}$ (3D translation + 6D rotation).
The \dy{five} baselines differ in how these conditions are injected and whether past/future contexts are used. Object motions are concatenated with human motions channel-wise. 

\paragraph{MDM~\cite{tevethuman}.}
A bidirectional transformer that denoises the current subsequence in one pass\dyt{, conditioned only on the last frame of the previous subsequence ($L^{\texttt{past}}{=}1$, $L^{\texttt{future}}{=}0$), analogous to the initial-pose condition used in short-term HOI synthesis}.
Each frame token is the linear projection of $[\vx_t, \vo^{L}_t, \vo^{R}_t]$.
All segment-level conditions are summed into a single per-segment token
$\vc_s = \tau(t) + \phi_\text{text}(\vw_s) + \phi_\text{scene}(\vs_s) + \phi_\text{bps}(\vg^{L}_s) + \phi_\text{bps}(\vg^{R}_s)$,
where $\tau(t)$ is the diffusion-timestep embedding and each $\phi$ is an MLP into $\mathbb{R}^{d_\text{model}}$.
A segment-restricted attention mask lets each frame attend only to its own segment's $\vc_s$.

\paragraph{CHOIS~\cite{li2024controllable}.}
Identical to MDM except that BPS is moved from the segment token to per-frame tokens, providing dense geometric guidance.
Frame tokens become $[\vx_t, \vo^{L}_t, \vo^{R}_t, \vg^{L}_{s(t)}, \vg^{R}_{s(t)}]$ where $s(t)$ is the segment containing frame $t$, and the segment token reduces to $\vc_s = \tau(t) + \phi_\text{text}(\vw_s) + \phi_\text{scene}(\vs_s)$.

\dy{\paragraph{DecHOI~\cite{jung2026decoupled}.}
Similar to CHOIS, DecHOI is designed for atomic HOI generation. Following its formulation, which takes object motion, BPS features, and a text embedding as input, we additionally provide the scene embedding alongside the text embedding for global conditioning. We also adopt the decoupled path denoiser and HOI adversarial discriminator from the original method for trajectory generation and interaction modeling, respectively, based on their official code repository. As DecHOI mainly tackles atomic motion generation, \dyt{we follow the same setting as MDM and CHOIS ($L^{\texttt{past}}{=}1$, $L^{\texttt{future}}{=}0$)}.}

\paragraph{InterDiff~\cite{xu2023interdiff}.}
An encoder--decoder transformer that uses $L^{\texttt{past}}{=}64$ past frames and $L^{\texttt{future}}{=}0$, matching its original prediction setting.
The encoder ingests the past frames, with each token aggregating a body-motion projection, per-slot object-motion projections gated by a hand-index embedding, the hand-specific BPS, and the past-segment text embedding.
The decoder denoises the body target over the past and current windows; each decoder token additionally incorporates per-frame BPS, per-frame text, the broadcast scene feature, and the timestep token, and attends to the encoder via standard cross-attention.

\paragraph{FlowMDM~\cite{barquero2024seamless}.}
A bidirectional denoiser that generates the full sequence in one pass.
FlowMDM replaces MDM's self-attention with the Pose-Centric Cross-Attention (PCCAT) block and its positional encoding with Blended Positional Encoding (BPE).
Conditioning enters as a per-frame vector
$\vc_t = \vw_{s(t)} + \vg^{L}_{s(t)} + \vg^{R}_{s(t)} + \vs_{s(t)} + \tau(t)$.
Inside each PCCAT layer, queries are formed from the concatenation of the motion feature and $\vc_t$, while keys and values come from the motion feature alone, so that the condition steers attention without entering the value space.

\dy{\subsection{Past and Future Conditioning for Baselines}}
\label{sec_supp:dagger}
\begin{table*}[t]
\caption{\textbf{\dy{Past and future conditioning for baselines}}}
\label{tab:sota_past_future}
\centering
\setlength{\tabcolsep}{3pt}
\resizebox{\textwidth}{!}{
\begin{tabular}{lcccccccccccccccccccc}
\toprule
\multirow{3}{*}{Method} & \multirow{3}{*}{\dy{$L^{\texttt{past}}$}} & \multirow{3}{*}{\dy{$L^{\texttt{future}}$}} &
\multicolumn{12}{c}{\textbf{Subsequence}} &
\multicolumn{3}{c}{\textbf{Transition}} &
\multicolumn{2}{c}{\textbf{Sequence}} \\
\cmidrule(lr){4-15}\cmidrule(lr){16-18}\cmidrule(lr){19-20}
& & &
\multicolumn{7}{c}{\textbf{Motion}} &
\multicolumn{5}{c}{\textbf{Contact}} &
\multicolumn{1}{c}{\textbf{Motion}} &
\multicolumn{2}{c}{\textbf{Jerk}} &
\multicolumn{2}{c}{\textbf{Scene}} \\
\cmidrule(lr){4-10}\cmidrule(lr){11-15}\cmidrule(lr){16-16}\cmidrule(lr){17-18}\cmidrule(lr){19-20}
& & &
MPJPE$\downarrow$ & H$_\mathrm{feet}\rightarrow$ & FS$\downarrow$ &
R@$3\uparrow$ & FID$\downarrow$ &
T$_\mathrm{goal}\downarrow$ & O$_\mathrm{goal}\downarrow$ &
C$_\mathrm{prec}\uparrow$ & C$_\mathrm{rec}\uparrow$ & C$_\mathrm{F1}\uparrow$ & C$_\%\rightarrow$ & P$_\mathrm{hand}\downarrow$ &
FID$\downarrow$ & PJ$\downarrow$ & AUJ$\downarrow$ &
Coll$_\%\downarrow$ & P$_\mathrm{max}\downarrow$ \\
\midrule
GT & - & - & - & -3.30 & 0.29 & - & - & - & - & - & - & - & 0.49 & 1.7e-6 & - & 0.03 & 2.22 & 0.35 & 0.04 \\
\midrule
MDM & \dy{1} & \dy{0} & \dyt{28.76} & \dyt{-6.39} & \dyt{0.33} & \dyt{0.43} & \dyt{0.70} & \dyt{54.14} & \upd{1.04} & \dyt{0.34} & \dyt{0.18} & \dyt{0.21} & \dyt{0.18} & \dyt{6.2e-5} & \dyt{1.59} & \dyt{0.16} & \dyt{2.05} & \dyt{0.67} & \dyt{0.07} \\
MDM$^\dagger$ & \dy{64} & \dy{64} & 22.50 & -4.82 & 0.30 & 0.49 & 0.45 & 31.47 & \upd{0.65} & 0.45 & 0.29 & 0.32 & 0.29 & 6.4e-5 & 0.70 & 0.10 & 2.06 & 0.52 & 0.06 \\
CHOIS & \dy{1} & \dy{0} & \dyt{28.47} & \dyt{-6.90} & \dyt{0.34} & \dyt{0.44} & \dyt{1.22} & \dyt{54.81} & \upd{1.07} & \dyt{0.33} & \dyt{0.18} & \dyt{0.20} & \dyt{0.17} & \dyt{3.4e-5} & \dyt{1.21} & \dyt{0.17} & \dyt{2.09} & \dyt{0.60} & \dyt{0.06} \\
CHOIS$^\dagger$ & \dy{64} & \dy{64} & 20.83 & -3.50 & 0.29 & 0.50 & 0.32 & 29.17 & \upd{0.58} & 0.44 & 0.30 & 0.32 & 0.29 & 8.7e-5 & 0.49 & 0.09 & 2.06 & 0.48 & 0.05 \\
\dy{DecHOI} & \dy{1} & \dy{0} & \upd{27.26} & \upd{-3.43} & \upd{0.39} & \dyt{0.44} & \dyt{0.53} & \upd{50.33} & \upd{0.98} & \upd{0.40} & \dyt{0.23} & \dyt{0.26} & \upd{0.24} & \upd{8.7e-5} & \dyt{2.44} & \upd{0.21} & \upd{2.23} & \upd{0.58} & \upd{0.06} \\
\dy{DecHOI$^\dagger$} & \dy{64} & \dy{64} & \dy{20.21} & \dy{-1.09} & \dy{0.40} & \dy{0.49} & \dy{0.62} & \dy{24.91} & \upd{0.51} & \dy{0.48} & \dy{0.37} & \dy{0.38} & \dy{0.37} & \dy{1.2e-4} & \dy{1.81} & \dy{0.13} & \dy{2.11} & \dy{0.49} & \dy{0.06} \\
InterDiff & \dy{64} & \dy{0} & 23.17 & -5.17 & 0.28 & 0.36 & 1.09 & 39.85 & 0.65 & 0.38 & 0.20 & 0.23 & 0.19 & 2.5e-6 & 4.60 & 0.80 & 3.44 & 0.51 & 0.06 \\
InterDiff$^\dagger$ & \dy{64} & \dy{64} & 19.84 & -3.86 & 0.31 & 0.49 & 0.18 & 28.03 & 0.43 & 0.40 & 0.25 & 0.27 & 0.24 & 4.5e-6 & 5.24 & 0.58 & 2.98 & 0.48 & 0.05 \\
\bottomrule
\end{tabular}
}
\end{table*}

\dy{
In Table~\ref{tab:sota}, each baseline follows its original temporal setting: MDM, CHOIS, and DecHOI are short-horizon models conditioned only on the last frame of the previous subsequence ($L^{\texttt{past}}{=}1$, $L^{\texttt{future}}{=}0$), and InterDiff forecasts future motion from past observations ($L^{\texttt{past}}{=}64$, $L^{\texttt{future}}{=}0$). To examine whether the gap stems from temporal context rather than architecture, we extend each baseline to take both past and future contexts with $L^{\texttt{past}}{=}L^{\texttt{future}}{=}64$, as in \ours, and denote the extended variants with $^\dagger$. All models are retrained from scratch on ParaHome for the same number of iterations, without using released weights; Table~\ref{tab:sota_past_future} lists the temporal context of each model. FlowMDM has no $\dagger$ variant as it generates the whole sequence jointly.}

\dy{
For MDM$^\dagger$, we construct segment-wise conditioning tokens for the past, current, and future subsequences from CLIP text features~\cite{radford2021learning}, BPS, and ViT scene features~\cite{dosovitskiy2021vit}, fuse them with the diffusion timestep embedding, and apply segment-wise masking in the self-attention layers so that each frame attends to the token of its own subsequence. CHOIS$^\dagger$ and DecHOI$^\dagger$ follow the same scheme, with tokens built from text and scene features only, since BPS is already provided per frame in these models. For InterDiff$^\dagger$, the encoder ingests the past motion as in the original model, while the decoder jointly predicts the past, current, and future motions.}

\dy{\noindent\textbf{Results.}
As shown in Table~\ref{tab:sota_past_future}, all four $\dagger$ variants improve over their original counterparts on most subsequence motion, contact, and transition metrics. Compared with a single past frame, longer past and future contexts reduce the goal-reaching error (\eg, 54.81~$\rightarrow$~29.17 for CHOIS and 54.14~$\rightarrow$~31.47 for MDM) and the transition FID (1.21~$\rightarrow$~0.49 for CHOIS and 1.59~$\rightarrow$~0.70 for MDM), indicating that the contexts help bridge adjacent subsequences. DecHOI$^\dagger$ likewise reduces MPJPE from 27.26 to 20.21, and InterDiff$^\dagger$, which already observes the past, further improves subsequence motion and contact quality with the added future context. These gains across architecturally distinct baselines suggest that past and future conditioning is a key factor for long-term HOI composition, beyond architectural differences.}

\dy{\subsection{Short-term HOI Synthesis of Baselines}
\label{sec_supp:atomic}}
\begin{table*}[t]
\caption{\textbf{Short-term HOI synthesis of baselines}}
\label{tab:supp_atomic}
\centering
\setlength{\tabcolsep}{3pt}
\resizebox{\textwidth}{!}{
\begin{tabular}{lcccccccccccc}
\toprule
Method & MPJPE$\downarrow$ & H$_\mathrm{feet}\rightarrow$ & FS$\downarrow$ & R@$3\uparrow$ & FID$\downarrow$ & T$_\mathrm{goal}\downarrow$ & O$_\mathrm{goal}\downarrow$ & C$_\mathrm{prec}\uparrow$ & C$_\mathrm{rec}\uparrow$ & C$_\mathrm{F1}\uparrow$ & C$_\%\rightarrow$ & P$_\mathrm{hand}\downarrow$ \\
\midrule
GT & -- & -3.66 & 0.28 & 0.49 & -- & -- & -- & -- & -- & -- & 0.70 & -- \\
\midrule
MDM & 12.63 & -4.40 & 0.24 & 0.50 & 0.26 & 17.59 & \upd{0.33} & 0.70 & 0.50 & 0.53 & 0.47 & 1.4e-4 \\
CHOIS & 12.37 & -4.30 & 0.23 & 0.49 & 0.22 & 16.94 & \upd{0.35} & 0.73 & 0.53 & 0.56 & 0.49 & 1.3e-4 \\
DecHOI & 12.04 & -4.26 & 0.27 & 0.50 & 0.22 & 15.86 & \upd{0.32} & 0.72 & 0.57 & 0.60 & 0.54 & 1.3e-4 \\
\bottomrule
\end{tabular}
}
\end{table*}

\dy{We evaluate the short-term HOI synthesis models (MDM, CHOIS, and DecHOI) on atomic HOI generation on ParaHome. Following the setup of short-term HOI synthesis~\cite{li2024controllable, jung2026decoupled}, we retrain these models with the ground-truth initial human pose as input; all other training inputs and model configurations are identical to Table~\ref{tab:sota}. Among the 989 test \dyb{subsequences} in ParaHome, 688 \dyb{subsequences} without transition motions are used in the evaluation. As shown in Table~\ref{tab:supp_atomic}, all baselines achieve results close to real motion on most metrics in the atomic HOI setting. Semantic correctness remains similar (R@3: 0.49--0.50 vs.\ 0.49), and low-level artifacts are also close to ground truth (foot height: $-4.26$ to $-4.40$\,cm vs.\ $-3.66$\,cm; foot sliding: 0.23--0.27 vs.\ 0.28). Although contact frequency is lower (0.47--0.54 vs.\ 0.70), the gap is modest compared with the degradation in long-term composition (MPJPE: \upd{27.26}\dyt{--28.76}\,cm, contact F1: \dyt{0.20--0.26} in Table~\ref{tab:sota}). These results suggest that their degradation reflects the difficulty of composition rather than an inability to synthesize atomic HOI on ParaHome or a domain gap with the dataset. Long-term composition requires maintaining transition continuity, anticipating future object interactions, and adapting to changing scene contexts---challenges that \ours~addresses through past and future conditioning.}

\section{Additional Experiments}
\label{sec_supp:exp}
In this section, we provide additional experimental results. Unless otherwise specified, we set the past context length $L^\texttt{past}$ and future context length $L^\texttt{future}$ to 64, and set the number of decoder layers $N^\texttt{dec}$ to 8.

\dy{\subsection{Ablation Studies}
\label{sec_supp:ablation}}
\noindent\textbf{Effect of $\lambda^\texttt{disp}$.}
\begin{table*}[t]
\caption{\textbf{Effect of $\lambda^\texttt{disp}$}}
\label{tab:supp_wdisp}
\centering
\setlength{\tabcolsep}{3pt}
\resizebox{\textwidth}{!}{
\begin{tabular}{ccccccccccccccccccc}
\toprule
\multirow{3}{*}{$\lambda^\texttt{disp}$} &
\multicolumn{12}{c}{\textbf{Subsequence}} &
\multicolumn{3}{c}{\textbf{Transition}} &
\multicolumn{2}{c}{\textbf{Sequence}} \\
\cmidrule(lr){2-13}\cmidrule(lr){14-16}\cmidrule(lr){17-18}
&
\multicolumn{7}{c}{\textbf{Motion}} &
\multicolumn{5}{c}{\textbf{Contact}} &
\multicolumn{1}{c}{\textbf{Motion}} &
\multicolumn{2}{c}{\textbf{Jerk}} &
\multicolumn{2}{c}{\textbf{Scene}} \\
\cmidrule(lr){2-8}\cmidrule(lr){9-13}\cmidrule(lr){14-14}\cmidrule(lr){15-16}\cmidrule(lr){17-18}
&
MPJPE$\downarrow$ & H$_\mathrm{feet}\rightarrow$ & FS$\downarrow$ &
R@$3\uparrow$ & FID$\downarrow$ &
T$_\mathrm{goal}\downarrow$ & O$_\mathrm{goal}\downarrow$ &
C$_\mathrm{prec}\uparrow$ & C$_\mathrm{rec}\uparrow$ & C$_\mathrm{F1}\uparrow$ & C$_\%\rightarrow$ & P$_\mathrm{hand}\downarrow$ &
FID$\downarrow$ & PJ$\downarrow$ & AUJ$\downarrow$ &
Coll$_\%\downarrow$ & P$_\mathrm{max}\downarrow$ \\
\midrule
GT & - & -3.30 & 0.29 & - & - & - & - & - & - & - & 0.49 & 1.7e-6 & - & 0.03 & 2.22 & 0.35 & 0.04 \\
\midrule
0.1 & 20.20 & -4.52 & 0.30 & 0.52 & 0.55 & 26.27 & \upd{0.52} & 0.46 & 0.35 & 0.37 & 0.34 & 8.5e-5 & 0.26 & 0.10 & 2.16 & 0.44 & 0.05 \\
0.2 & 19.77 & -4.08 & 0.32 & 0.53 & 0.34 & 25.07 & \upd{0.52} & 0.47 & 0.35 & 0.37 & 0.34 & 6.9e-5 & 0.23 & 0.11 & 2.16 & 0.45 & 0.05 \\
0.3 & 19.85 & -4.47 & 0.35 & 0.54 & 0.41 & 25.70 & \upd{0.52} & 0.46 & 0.35 & 0.36 & 0.34 & 8.6e-5 & 0.81 & 0.13 & 2.15 & 0.46 & 0.05 \\
0.4 & 19.57 & -4.84 & 0.34 & 0.54 & 0.27 & 25.02 & \upd{0.51} & 0.47 & 0.37 & 0.38 & 0.35 & 8.0e-5 & 0.46 & 0.14 & 2.17 & 0.46 & 0.05 \\
0.5 & 19.92 & -4.90 & 0.35 & 0.50 & 0.39 & 25.75 & \upd{0.53} & 0.47 & 0.37 & 0.38 & 0.36 & 7.9e-5 & 0.75 & 0.15 & 2.19 & 0.46 & 0.05 \\
\bottomrule
\end{tabular}
}
\end{table*}

To analyze the effect of the hand--object displacement loss weight $\lambda^\texttt{disp}$, we vary its value from 0.1 to 0.5. As shown in Table~\ref{tab:supp_wdisp}, increasing $\lambda^\texttt{disp}$ generally improves contact metrics, indicating that the displacement loss encourages more accurate hand--object interactions. However, larger values slightly degrade motion quality. Balancing contact performance and motion quality, we set $\lambda^\texttt{disp} = 0.3$ and fix this value for all subsequent experiments.

\noindent\textbf{Self-attention vs.\ cross-attention.}
\begin{table*}[!t]
\caption{\textbf{Self-attention vs.\ cross-attention}}
\label{tab:supp_arch}
\centering
\setlength{\tabcolsep}{3pt}
\resizebox{\textwidth}{!}{
\begin{tabular}{lcccccccccccccccccc}
\toprule
\multirow{3}{*}{Attn type} &
\multicolumn{12}{c}{\textbf{Subsequence}} &
\multicolumn{3}{c}{\textbf{Transition}} &
\multicolumn{2}{c}{\textbf{Sequence}} \\
\cmidrule(lr){2-13}\cmidrule(lr){14-16}\cmidrule(lr){17-18}
&
\multicolumn{7}{c}{\textbf{Motion}} &
\multicolumn{5}{c}{\textbf{Contact}} &
\multicolumn{1}{c}{\textbf{Motion}} &
\multicolumn{2}{c}{\textbf{Jerk}} &
\multicolumn{2}{c}{\textbf{Scene}} \\
\cmidrule(lr){2-8}\cmidrule(lr){9-13}\cmidrule(lr){14-14}\cmidrule(lr){15-16}\cmidrule(lr){17-18}
&
MPJPE$\downarrow$ & H$_\mathrm{feet}\rightarrow$ & FS$\downarrow$ &
R@$3\uparrow$ & FID$\downarrow$ &
T$_\mathrm{goal}\downarrow$ & O$_\mathrm{goal}\downarrow$ &
C$_\mathrm{prec}\uparrow$ & C$_\mathrm{rec}\uparrow$ & C$_\mathrm{F1}\uparrow$ & C$_\%\rightarrow$ & P$_\mathrm{hand}\downarrow$ &
FID$\downarrow$ & PJ$\downarrow$ & AUJ$\downarrow$ &
Coll$_\%\downarrow$ & P$_\mathrm{max}\downarrow$ \\
\midrule
GT & - & -3.30 & 0.29 & - & - & - & - & - & - & - & 0.49 & 1.7e-6 & - & 0.03 & 2.22 & 0.35 & 0.04 \\
\midrule
Self-attn & 19.90 & -5.49 & 0.39 & 0.52 & 0.49 & 25.73 & \upd{0.51} & 0.47 & 0.36 & 0.37 & 0.35 & 8.8e-5 & 1.44 & 0.20 & 2.20 & 0.46 & 0.05 \\
Cross-attn & 19.85 & -4.47 & 0.35 & 0.54 & 0.41 & 25.70 & \upd{0.52} & 0.46 & 0.35 & 0.36 & 0.34 & 8.6e-5 & 0.81 & 0.13 & 2.15 & 0.46 & 0.05 \\
\bottomrule
\end{tabular}
}
\end{table*}

To analyze the effect of the attention strategy for multimodal conditioning, we compare our cross-attention design with a variant that replaces cross-attention layers with self-attention. In this variant, all multimodal conditions---scene, object geometry, and text features---are concatenated to the input sequence of the denoising transformer $f$, analogous to the diffusion timestep embedding shown in Fig.~\ref{fig:pipeline}. As shown in Table~\ref{tab:supp_arch}, \dyb{cross-attention improves motion quality and transition smoothness (\eg, transition FID 1.44 $\rightarrow$ 0.81, PJ 0.20 $\rightarrow$ 0.13), while contact metrics remain comparable.} We hypothesize that cross-attention decouples the motion and conditioning representation spaces: motion tokens attend to conditioning signals without the conditioning features being modified in return, allowing the model to preserve task-relevant motion structure while incorporating external guidance.

\begin{table*}[!t]
\caption{\textbf{Effect of the number of decoder layers $N^\texttt{dec}$}}
\label{tab:supp_ndec}
\centering
\setlength{\tabcolsep}{3pt}
\resizebox{\textwidth}{!}{
\begin{tabular}{ccccccccccccccccccc}
\toprule
\multirow{3}{*}{$N^\texttt{dec}$} &
\multicolumn{12}{c}{\textbf{Subsequence}} &
\multicolumn{3}{c}{\textbf{Transition}} &
\multicolumn{2}{c}{\textbf{Sequence}} \\
\cmidrule(lr){2-13}\cmidrule(lr){14-16}\cmidrule(lr){17-18}
&
\multicolumn{7}{c}{\textbf{Motion}} &
\multicolumn{5}{c}{\textbf{Contact}} &
\multicolumn{1}{c}{\textbf{Motion}} &
\multicolumn{2}{c}{\textbf{Jerk}} &
\multicolumn{2}{c}{\textbf{Scene}} \\
\cmidrule(lr){2-8}\cmidrule(lr){9-13}\cmidrule(lr){14-14}\cmidrule(lr){15-16}\cmidrule(lr){17-18}
&
MPJPE$\downarrow$ & H$_\mathrm{feet}\rightarrow$ & FS$\downarrow$ &
R@$3\uparrow$ & FID$\downarrow$ &
T$_\mathrm{goal}\downarrow$ & O$_\mathrm{goal}\downarrow$ &
C$_\mathrm{prec}\uparrow$ & C$_\mathrm{rec}\uparrow$ & C$_\mathrm{F1}\uparrow$ & C$_\%\rightarrow$ & P$_\mathrm{hand}\downarrow$ &
FID$\downarrow$ & PJ$\downarrow$ & AUJ$\downarrow$ &
Coll$_\%\downarrow$ & P$_\mathrm{max}\downarrow$ \\
\midrule
GT & - & -3.30 & 0.29 & - & - & - & - & - & - & - & 0.49 & 1.7e-6 & - & 0.03 & 2.22 & 0.35 & 0.04 \\
\midrule
6 & 20.11 & -6.04 & 0.36 & 0.54 & 0.43 & 25.64 & \upd{0.53} & 0.48 & 0.37 & 0.38 & 0.36 & 1.1e-4 & 0.54 & 0.11 & 2.16 & 0.45 & 0.05 \\
8 & 19.85 & -4.47 & 0.35 & 0.54 & 0.41 & 25.70 & \upd{0.52} & 0.46 & 0.35 & 0.36 & 0.34 & 8.6e-5 & 0.81 & 0.13 & 2.15 & 0.46 & 0.05 \\
10 & 19.90 & -4.73 & 0.34 & 0.55 & 0.43 & 25.84 & \upd{0.55} & 0.46 & 0.35 & 0.36 & 0.34 & 1.0e-4 & 0.54 & 0.15 & 2.15 & 0.47 & 0.05 \\
\bottomrule
\end{tabular}
}
\end{table*}

\begin{table*}[t!]
\caption{\textbf{Numerical results of Fig.~\ref{fig:context_ablation}}}
\label{tab:context}
\centering
\setlength{\tabcolsep}{3pt}
\resizebox{\textwidth}{!}{
\begin{tabular}{lccccccccccccccccccc}
\toprule
\multirow{3}{*}{method} &
\multirow{3}{*}{\makecell{$L^\texttt{past}$}} &
\multirow{3}{*}{\makecell{$L^\texttt{future}$}} &
\multicolumn{12}{c}{\textbf{Subsequence}} &
\multicolumn{3}{c}{\textbf{Transition}} &
\multicolumn{2}{c}{\textbf{Sequence}} \\
\cmidrule(lr){4-15}\cmidrule(lr){16-18}\cmidrule(lr){19-20}
& & &
\multicolumn{7}{c}{\textbf{Motion}} &
\multicolumn{5}{c}{\textbf{Contact}} &
\multicolumn{1}{c}{\textbf{Motion}} &
\multicolumn{2}{c}{\textbf{Jerk}} &
\multicolumn{2}{c}{\textbf{Scene}} \\
\cmidrule(lr){4-10}\cmidrule(lr){11-15}\cmidrule(lr){16-16}\cmidrule(lr){17-18}\cmidrule(lr){19-20}
& & &
MPJPE$\downarrow$ & H$_\mathrm{feet}\rightarrow$ & FS$\downarrow$ &
R@$3\uparrow$ & FID$\downarrow$ &
T$_\mathrm{goal}\downarrow$ & O$_\mathrm{goal}\downarrow$ &
C$_\mathrm{prec}\uparrow$ & C$_\mathrm{rec}\uparrow$ & C$_\mathrm{F1}\uparrow$ & C$_\%\rightarrow$ & P$_\mathrm{hand}\downarrow$ &
FID$\downarrow$ & PJ$\downarrow$ & AUJ$\downarrow$ &
Coll$_\%\downarrow$ & P$_\mathrm{max}\downarrow$ \\
\midrule
GT & - & - & - & -3.30 & 0.29 & - & - & - & - & - & - & - & 0.49 & 1.7e-6 & - & 0.03 & 2.22 & 0.35 & 0.04 \\
\midrule
\multicolumn{20}{l}{\textit{Effect of past context length ($L^\texttt{future}=64$)}} \\
\ours & 0 & 64 & 19.69 & -0.14 & 0.32 & 0.51 & 0.46 & 34.95 & \upd{0.51} & 0.32 & 0.22 & 0.23 & 0.21 & 6.2e-5 & 20.72 & 0.58 & 2.92 & 0.42 & 0.05 \\
 & 1 & 64 & 20.18 & -4.83 & 0.31 & 0.52 & 0.36 & 26.71 & \upd{0.55} & 0.46 & 0.33 & 0.35 & 0.32 & 8.5e-5 & 1.15 & 0.19 & 2.13 & 0.47 & 0.05 \\
 & 32 & 64 & 19.53 & -3.98 & 0.34 & 0.53 & 0.31 & 25.14 & \upd{0.51} & 0.47 & 0.36 & 0.38 & 0.35 & 7.0e-5 & 0.63 & 0.15 & 2.16 & 0.43 & 0.05 \\
 & 64 & 64 & 19.85 & -4.47 & 0.35 & 0.54 & 0.41 & 25.70 & \upd{0.52} & 0.46 & 0.35 & 0.36 & 0.34 & 8.6e-5 & 0.81 & 0.13 & 2.15 & 0.45 & 0.05 \\
 & 128 & 64 & 18.89 & -4.44 & 0.33 & 0.52 & 0.42 & 23.72 & \upd{0.48} & 0.48 & 0.37 & 0.38 & 0.35 & 7.9e-5 & 0.25 & 0.12 & 2.15 & 0.44 & 0.05 \\
 & 256 & 64 & 18.80 & -4.54 & 0.31 & 0.53 & 0.53 & 23.46 & \upd{0.47} & 0.47 & 0.36 & 0.38 & 0.36 & 7.2e-5 & 0.22 & 0.10 & 2.15 & 0.44 & 0.05 \\
\midrule
\multicolumn{20}{l}{\textit{Effect of future context length ($L^\texttt{past}=64$)}} \\
\ours & 64 & 0 & 27.40 & -7.04 & 0.33 & 0.43 & 0.88 & 54.25 & \upd{0.99} & 0.34 & 0.20 & 0.22 & 0.20 & 3.5e-5 & 1.42 & 0.17 & 2.19 & 0.56 & 0.06 \\
 & 64 & 1 & 23.47 & -7.30 & 0.33 & 0.47 & 0.70 & 36.27 & \upd{0.70} & 0.42 & 0.29 & 0.31 & 0.29 & 7.1e-5 & 1.76 & 0.14 & 2.17 & 0.54 & 0.06 \\
 & 64 & 32 & 20.96 & -3.03 & 0.34 & 0.52 & 0.22 & 28.49 & \upd{0.59} & 0.45 & 0.33 & 0.34 & 0.32 & 7.5e-5 & 0.86 & 0.13 & 2.16 & 0.47 & 0.05 \\
 & 64 & 64 & 19.85 & -4.47 & 0.35 & 0.54 & 0.41 & 25.70 & \upd{0.52} & 0.46 & 0.35 & 0.36 & 0.34 & 8.6e-5 & 0.81 & 0.13 & 2.15 & 0.45 & 0.05 \\
 & 64 & 128 & 19.39 & -6.75 & 0.33 & 0.53 & 0.58 & 25.13 & \upd{0.50} & 0.47 & 0.37 & 0.38 & 0.35 & 1.1e-4 & 0.24 & 0.12 & 2.16 & 0.47 & 0.05 \\
 & 64 & 256 & 18.11 & -5.25 & 0.34 & 0.53 & 0.53 & 21.07 & \upd{0.41} & 0.48 & 0.37 & 0.39 & 0.36 & 7.9e-5 & 0.22 & 0.12 & 2.14 & 0.43 & 0.05 \\
\bottomrule
\end{tabular}
}
\end{table*}

\noindent\textbf{Effect of the number of decoder layers $N^\texttt{dec}$.}
We conduct an ablation study by varying the number of decoder layers $N^\texttt{dec}$ from 6 to 10, as shown in Table~\ref{tab:supp_ndec}. Performance varies modestly across depths, and the improvements from 8 to 10 layers are marginal compared to the increased computational cost. Therefore, we adopt $N^\texttt{dec} = 8$ as the default configuration to balance performance and efficiency.

\noindent\textbf{Numerical results of Fig.~\ref{fig:context_ablation}.}
Full results of Fig.~\ref{fig:context_ablation}, \dy{\ie, the effect of past and future context lengths,} are presented in Table~\ref{tab:context}.

\dy{\subsection{Scene Representation}
\label{sec_supp:scene}}
\begin{table*}[t!]
\caption{\textbf{Effect of scene encoder}}
\label{tab:scene_encoder}
\centering
\setlength{\tabcolsep}{3pt}
\resizebox{\textwidth}{!}{
\begin{tabular}{lcccccccccccccccccc}
\toprule
\multirow{3}{*}{method} &
\multicolumn{12}{c}{\textbf{Subsequence}} &
\multicolumn{3}{c}{\textbf{Transition}} &
\multicolumn{2}{c}{\textbf{Sequence}} \\
\cmidrule(lr){2-13}\cmidrule(lr){14-16}\cmidrule(lr){17-18}
&
\multicolumn{7}{c}{\textbf{Motion}} &
\multicolumn{5}{c}{\textbf{Contact}} &
\multicolumn{1}{c}{\textbf{Motion}} &
\multicolumn{2}{c}{\textbf{Jerk}} &
\multicolumn{2}{c}{\textbf{Scene}} \\
\cmidrule(lr){2-8}\cmidrule(lr){9-13}\cmidrule(lr){14-14}\cmidrule(lr){15-16}\cmidrule(lr){17-18}
&
MPJPE$\downarrow$ & H$_\mathrm{feet}\rightarrow$ & FS$\downarrow$ &
R@$3\uparrow$ & FID$\downarrow$ &
T$_\mathrm{goal}\downarrow$ & O$_\mathrm{goal}\downarrow$ &
C$_\mathrm{prec}\uparrow$ & C$_\mathrm{rec}\uparrow$ & C$_\mathrm{F1}\uparrow$ & C$_\%\rightarrow$ & P$_\mathrm{hand}\downarrow$ &
FID$\downarrow$ & PJ$\downarrow$ & AUJ$\downarrow$ &
Coll$_\%\downarrow$ & P$_\mathrm{max}\downarrow$ \\
\midrule
GT & - & -3.30 & 0.29 & - & - & - & - & - & - & - & 0.49 & 1.7e-6 & - & 0.03 & 2.22 & 0.35 & 0.04 \\
\midrule
CNN & 19.65 & -5.94 & 0.34 & 0.52 & 0.74 & 24.95 & \upd{0.50} & 0.47 & 0.36 & 0.37 & 0.35 & 8.9e-5 & 0.61 & 0.13 & 2.15 & 0.45 & 0.05 \\
ViT & 19.85 & -4.47 & 0.35 & 0.54 & 0.41 & 25.70 & \upd{0.52} & 0.46 & 0.35 & 0.36 & 0.34 & 8.6e-5 & 0.81 & 0.13 & 2.15 & 0.46 & 0.05 \\
\bottomrule
\end{tabular}
}
\end{table*}

\begin{table*}[t]
\caption{\textbf{2D BEV vs.\ 3D voxel scene representation}}
\label{tab:supp_voxel}
\centering
\setlength{\tabcolsep}{3pt}
\resizebox{\textwidth}{!}{
\begin{tabular}{lccccccccccccccccc}
\toprule
Scene & MPJPE$\downarrow$ & H$_\mathrm{feet}\rightarrow$ & FS$\downarrow$ & R@$3\uparrow$ & FID$\downarrow$ & T$_\mathrm{goal}\downarrow$ & O$_\mathrm{goal}\downarrow$ & C$_\mathrm{prec}\uparrow$ & C$_\mathrm{rec}\uparrow$ & C$_\mathrm{F1}\uparrow$ & C$_\%\rightarrow$ & P$_\mathrm{hand}\downarrow$ & FID$_\mathrm{tr}\downarrow$ & PJ$\downarrow$ & AUJ$\downarrow$ & Coll$_\%\downarrow$ & P$_\mathrm{max}\downarrow$ \\
\midrule
GT & -- & -3.30 & 0.29 & 0.49 & -- & -- & -- & -- & -- & -- & 0.49 & 1.7e-6 & -- & 0.03 & 2.22 & 0.35 & 0.04 \\
\midrule
2D BEV & 19.85 & -4.47 & 0.35 & 0.54 & 0.41 & 25.70 & \upd{0.52} & 0.46 & 0.35 & 0.36 & 0.34 & 8.6e-5 & 0.81 & 0.13 & 2.15 & 0.45 & 0.05 \\
3D voxel & 19.16 & -4.34 & 0.33 & 0.52 & 0.48 & 24.95 & \upd{0.51} & 0.47 & 0.37 & 0.38 & 0.36 & 7.4e-5 & 0.34 & 0.13 & 2.16 & 0.44 & 0.05 \\
\bottomrule
\end{tabular}
}
\end{table*}

\noindent\dy{\textbf{Effect of scene encoder.}}
To verify the effect of the scene encoder architecture, we compare CNN and ViT as the scene encoder backbone in Table~\ref{tab:scene_encoder}. CNN captures local spatial features through hierarchical convolutions, while ViT models global spatial context via self-attention over image patches. In this experiment, we use a 4-layer CNN. Although CNN yields marginally better motion quality---likely due to its locality bias being well-suited for processing the structured 2D scene layout---both variants achieve comparable performance across most metrics. The CNN encoder achieves slightly lower MPJPE (19.65 vs.\ 19.85) and better $\text{T}_{\text{goal}}$ (24.95 vs.\ 25.70), while ViT achieves better subsequence FID (0.41 vs.\ 0.74). These differences are minor, suggesting that our framework is robust to the choice of scene encoder architecture. We adopt ViT as the default due to its flexibility and scalability.

\noindent\dy{\textbf{2D BEV vs.\ 3D voxel scene representation.}}
\label{sec_supp:voxel}
\dy{To evaluate whether a 3D scene representation benefits full-body HOI generation, we replace our BEV scene representation with a 3D voxel. Specifically, we voxelize the full scene at 2\,cm resolution, extract a human-centric $32^3$ occupancy volume aligned with the human heading direction, and encode it using a ViT following TRUMANS~\cite{jiang2024scaling}. For a clear comparison, DNO optimization is disabled in both settings. As shown in Table~\ref{tab:supp_voxel}, adopting a 3D voxel representation improves MPJPE (19.85 $\rightarrow$ 19.16\,cm) and contact F1 (0.36 $\rightarrow$ 0.38), with a slight reduction in collision ratio (Coll$_\%$: 0.45 $\rightarrow$ 0.44). However, these gains come with a substantial computational overhead: the voxel variant requires approximately $4\times$ longer training due to the full-scene 3D I/O pipeline. We therefore find that the 2D BEV representation offers a more practical balance between geometric awareness and efficiency.}

\dy{\subsection{Robustness to Object Motion Perturbations}
\label{sec_supp:robustness}}
\begin{table*}[t]
\caption{\textbf{Robustness to object motion perturbations}}
\label{tab:supp_robustness}
\centering
\setlength{\tabcolsep}{3pt}
\resizebox{\textwidth}{!}{
\begin{tabular}{cccccccccccccccccccc}
\toprule
$\sigma$ (cm) & $p$ (\%) & T$_\mathrm{obj}$ (cm) & MPJPE$\downarrow$ & H$_\mathrm{feet}\rightarrow$ & FS$\downarrow$ & R@$3\uparrow$ & FID$\downarrow$ & T$_\mathrm{goal}\downarrow$ & O$_\mathrm{goal}\downarrow$ & C$_\mathrm{prec}\uparrow$ & C$_\mathrm{rec}\uparrow$ & C$_\mathrm{F1}\uparrow$ & C$_\%\rightarrow$ & P$_\mathrm{hand}\downarrow$ & FID$_\mathrm{tr}\downarrow$ & PJ$\downarrow$ & AUJ$\downarrow$ & Coll$_\%\downarrow$ & P$_\mathrm{max}\downarrow$ \\
\midrule
GT & -- & -- & - & -3.30 & 0.29 & 0.49 & -- & -- & -- & -- & -- & -- & 0.49 & 1.7e-6 & -- & 0.03 & 2.22 & 0.35 & 0.04 \\
\midrule
0 & 0 & 0.00 & 19.85 & -4.47 & 0.35 & 0.54 & 0.41 & 25.70 & \upd{0.52} & 0.46 & 0.35 & 0.36 & 0.34 & 8.6e-5 & 0.81 & 0.13 & 2.15 & 0.45 & 0.05 \\
\midrule
\multicolumn{20}{l}{\textit{(A) Temporal coverage}} \\
5 & 10 & 0.55 & 19.80 & -4.26 & 0.34 & 0.50 & 0.44 & 25.70 & \upd{0.53} & 0.46 & 0.36 & 0.37 & 0.35 & 9.5e-5 & 0.72 & 0.12 & 2.16 & 0.46 & 0.05 \\
5 & 25 & 1.33 & 19.93 & -3.77 & 0.34 & 0.51 & 0.43 & 25.84 & \upd{0.53} & 0.46 & 0.34 & 0.36 & 0.34 & 1.0e-4 & 0.77 & 0.13 & 2.16 & 0.46 & 0.05 \\
5 & 50 & 2.68 & 19.98 & -4.42 & 0.34 & 0.52 & 0.43 & 26.09 & \upd{0.53} & 0.46 & 0.34 & 0.36 & 0.34 & 9.3e-5 & 0.66 & 0.13 & 2.16 & 0.47 & 0.05 \\
5 & 75 & 4.19 & 19.91 & -4.19 & 0.33 & 0.53 & 0.42 & 26.42 & \upd{0.53} & 0.46 & 0.33 & 0.36 & 0.32 & 9.2e-5 & 1.17 & 0.13 & 2.16 & 0.46 & 0.05 \\
5 & 100 & 5.33 & 19.80 & -4.72 & 0.34 & 0.53 & 0.43 & 25.23 & \upd{0.51} & 0.45 & 0.34 & 0.35 & 0.34 & 8.8e-5 & 1.00 & 0.13 & 2.16 & 0.47 & 0.05 \\
\midrule
\multicolumn{20}{l}{\textit{(B) Noise magnitude}} \\
0.5 & 100 & 0.53 & 19.71 & -4.03 & 0.34 & 0.53 & 0.41 & 25.64 & \upd{0.52} & 0.46 & 0.35 & 0.36 & 0.35 & 9.6e-5 & 0.82 & 0.13 & 2.16 & 0.46 & 0.05 \\
1 & 100 & 1.07 & 19.74 & -4.22 & 0.35 & 0.52 & 0.39 & 25.20 & \upd{0.52} & 0.46 & 0.35 & 0.37 & 0.35 & 1.1e-4 & 0.87 & 0.13 & 2.16 & 0.47 & 0.05 \\
2 & 100 & 2.13 & 19.62 & -4.17 & 0.33 & 0.53 & 0.39 & 25.35 & \upd{0.52} & 0.46 & 0.35 & 0.36 & 0.34 & 9.5e-5 & 1.18 & 0.13 & 2.16 & 0.47 & 0.05 \\
5 & 100 & 5.33 & 19.80 & -4.72 & 0.34 & 0.53 & 0.43 & 25.23 & \upd{0.51} & 0.45 & 0.34 & 0.35 & 0.34 & 8.8e-5 & 1.00 & 0.13 & 2.16 & 0.47 & 0.05 \\
10 & 100 & 10.67 & 20.28 & -4.32 & 0.34 & 0.51 & 0.48 & 26.59 & \upd{0.54} & 0.46 & 0.29 & 0.33 & 0.29 & 7.7e-5 & 0.98 & 0.13 & 2.16 & 0.48 & 0.06 \\
\bottomrule
\end{tabular}
}
\end{table*}

\dy{To evaluate robustness to imperfect object motion inputs, we perturb the object trajectories at inference time along two axes: (A) temporal coverage and (B) noise magnitude. Specifically, we perturb object translations using temporally smooth random 3D offsets generated by low-pass filtering of random noise, which better simulates realistic trajectory errors than frame-level jitter. To isolate the effect of noisy object motion conditioning, DNO optimization is disabled in this experiment. We additionally report the mean object translation error T$_\mathrm{obj}$ to quantify the actual deviation from the original object motion. In Table~\ref{tab:supp_robustness}, $\sigma$ denotes the noise magnitude and $p$ the perturbed portion of the trajectory.

\textbf{Temporal coverage:} We gradually increase the portion of the trajectory affected by perturbation while fixing the noise magnitude to $\sigma=5$\,cm. \ours~demonstrates stable performance against temporal corruption. Even as we increase the corrupted portion of the trajectory up to 100\%, the overall metrics remain comparable to the clean setting. 

\textbf{Noise magnitude:} We vary the perturbation magnitude from $\sigma=0.5$\,cm to $\sigma=10$\,cm, applying noise to the entire trajectory. \ours~remains stable up to $\sigma=5$\,cm, maintaining a contact F1 of 0.35. Even under the perturbation of $\sigma=10$\,cm, while contact recall decreases, the precision remains unaffected. These results suggest that \ours~remains effective under moderate object motion errors, beyond the ideal ground-truth object motion setting.}

\dy{\subsection{DNO Analysis}
\label{sec_supp:dno}}
\begin{table}[t]
\caption{\textbf{Per-body-part collision analysis}}
\label{tab:supp_dno_part}
\centering
\setlength{\tabcolsep}{3pt}
\small
\begin{tabular}{lcccccc}
\toprule
& \multicolumn{3}{c}{(A) All frames} & \multicolumn{3}{c}{(B) Pelvis cleared by DNO} \\
\cmidrule(lr){2-4}\cmidrule(lr){5-7}
Part & GT & w/o DNO & w/ DNO & GT & w/o DNO & w/ DNO \\
\midrule
hand & 0.271 & 0.278 & 0.250 & 0.373 & 0.450 & 0.348 \\
arm & 0.091 & 0.150 & 0.104 & 0.254 & 0.422 & 0.233 \\
leg & 0.056 & 0.157 & 0.031 & 0.178 & 0.627 & 0.102 \\
torso & 0.051 & 0.102 & 0.015 & 0.229 & 0.470 & 0.092 \\
pelvis & 0.044 & 0.106 & 0.024 & 0.167 & 0.429 & 0.084 \\
head & 0.027 & 0.032 & 0.028 & 0.025 & 0.069 & 0.034 \\
foot & 0.004 & 0.001 & 0.001 & -- & -- & -- \\
any (= Coll$_\%$) & 0.357 & 0.454 & 0.314 & 0.623 & 0.997 & 0.519 \\
\bottomrule
\end{tabular}
\end{table}

\begin{table*}[t]
\caption{\textbf{Extending the DNO objective to all joints}}
\label{tab:supp_dno_objective}
\centering
\setlength{\tabcolsep}{3pt}
\resizebox{\textwidth}{!}{
\begin{tabular}{lccccccccccccccccc}
\toprule
\multirow{3}{*}{DNO objective} &
\multicolumn{12}{c}{\textbf{Subsequence}} &
\multicolumn{3}{c}{\textbf{Transition}} &
\multicolumn{2}{c}{\textbf{Sequence}} \\
\cmidrule(lr){2-13}\cmidrule(lr){14-16}\cmidrule(lr){17-18}
&
\multicolumn{7}{c}{\textbf{Motion}} &
\multicolumn{5}{c}{\textbf{Contact}} &
\multicolumn{1}{c}{\textbf{Motion}} &
\multicolumn{2}{c}{\textbf{Jerk}} &
\multicolumn{2}{c}{\textbf{Scene}} \\
\cmidrule(lr){2-8}\cmidrule(lr){9-13}\cmidrule(lr){14-14}\cmidrule(lr){15-16}\cmidrule(lr){17-18}
&
MPJPE$\downarrow$ & H$_\mathrm{feet}\rightarrow$ & FS$\downarrow$ &
R@$3\uparrow$ & FID$\downarrow$ &
T$_\mathrm{goal}\downarrow$ & O$_\mathrm{goal}\downarrow$ &
C$_\mathrm{prec}\uparrow$ & C$_\mathrm{rec}\uparrow$ & C$_\mathrm{F1}\uparrow$ & C$_\%\rightarrow$ & P$_\mathrm{hand}\downarrow$ &
FID$\downarrow$ & PJ$\downarrow$ & AUJ$\downarrow$ &
Coll$_\%\downarrow$ & P$_\mathrm{max}\downarrow$ \\
\midrule
pelvis only (ours) & 19.10 & -4.52 & 0.33 & 0.52 & 0.28 & 24.64 & 0.47 & 0.47 & 0.37 & 0.38 & 0.35 & 6.7e-5 & 1.07 & 0.13 & 2.12 & 0.33 & 0.05 \\
per-joint & 19.21 & -4.04 & 0.34 & 0.52 & 0.19 & 25.46 & 0.50 & 0.46 & 0.32 & 0.34 & 0.31 & 6.0e-5 & 1.07 & 0.13 & 2.13 & 0.31 & 0.04 \\
\bottomrule
\end{tabular}
}
\end{table*}

\dy{\textbf{What DNO optimizes and why collisions remain.}
The collision objective in DNO and the reported collision metric measure different aspects of collision. The DNO objective is designed to correct coarse navigation-level conflicts by optimizing the pelvis trajectory on the ground plane. In contrast, Coll$_\%$ reports the fraction of frames where any body vertex penetrates the scene geometry by more than 3\,cm, over all body parts. Under this full-body metric, DNO reduces Coll$_\%$ to a level comparable to the ground-truth motion collision ratio of 0.35. In the GT motion, approximately 90\% of the intersections belong to the hands (65.4\%) and head (24.5\%). Since the evaluation metric uses a strict 3\,cm penetration threshold, it also captures genuine close-range contacts (\eg, hand-object interactions), and the non-zero collision ratio of GT motion suggests that this metric has a non-negligible floor on contact-rich sequences. On a pelvis-level collision metric that directly evaluates DNO on the axis it optimizes, \ie, the fraction of frames in which the predicted pelvis projected onto the ground plane lies in an occupied region of the BEV scene map, DNO reduces collisions from 0.09 to 0.01, an 89\% reduction.}

\dy{\textbf{Per-body-part collision analysis.}
Table~\ref{tab:supp_dno_part}(A) reports the per-part penetration ratio over all test frames (989 \dyb{subsequences}, 95,703 frames), \ie, the fraction of frames containing at least one vertex of each body part penetrating the scene by more than 3\,cm, using the same 3D SDF query as Coll$_\%$; the union over all parts corresponds to Coll$_\%$. Before DNO, leg penetration is the main source of intersection (15.7\% vs.\ 5.6\% in GT), mainly near dining tables. Hand penetration remains close to GT, reflecting that the metric evaluates rigid meshes against a dilated voxel SDF, where even captured motion has non-zero collisions. Although DNO optimizes only the pelvis trajectory, it reduces collisions in other body parts through the kinematic chain: relative to w/o DNO, penetration decreases by 0.125 for legs, 0.046 for arms, and 0.028 for hands. After DNO, five of seven body parts have lower penetration than GT, while \dyb{the arm (+0.013) and head (+0.001) are slightly higher}.
We further isolate frames where the pelvis initially intersected the 2D obstacle region but became collision-free after DNO (12,670 frames, 13.2\% of the test set); DNO moved the pelvis outside the obstacle region in 95.7\% of these frames. As shown in Table~\ref{tab:supp_dno_part}(B), these frames correspond to close-proximity interactions, where captured motion already has a high collision rate (62.3\%). Without DNO, 99.7\% of frames contain penetration, while DNO reduces penetration across all body parts toward GT levels, and the residual collisions remain comparable to captured motion. Contact F1 is unchanged, and hand joint error improves by 16.0\,mm.}

\dy{\textbf{Extending the DNO objective to all joints.}
We implement a per-joint collision objective and compare it with our pelvis-level objective on the full test set (Table~\ref{tab:supp_dno_objective}); both use 25 DDIM steps and 100 optimization steps. The per-joint objective further reduces collisions (0.33 $\rightarrow$ 0.31) and penetration depth (0.05 $\rightarrow$ 0.04), but this comes at the cost of interaction quality: contact F1 decreases from 0.38 to 0.34, mainly due to reduced recall (0.37 $\rightarrow$ 0.32), and T$_\mathrm{goal}$ increases from 24.64 to 25.46. Therefore, although per-joint optimization achieves stronger collision reduction, we retain the pelvis-level objective as it provides a better balance between collision avoidance and functional interaction quality.}

\begin{table*}[t]
\caption{\textbf{DNO configuration and runtime}}
\label{tab:supp_dno_runtime}
\centering
\setlength{\tabcolsep}{3pt}
\resizebox{\textwidth}{!}{
\begin{tabular}{lccccccccccccccccccccc}
\toprule
\multirow{3}{*}{Model} & \multirow{3}{*}{\shortstack{DDIM\\steps}} & \multirow{3}{*}{\shortstack{Opt.\\steps}} &
\multicolumn{12}{c}{\textbf{Subsequence}} &
\multicolumn{3}{c}{\textbf{Transition}} &
\multicolumn{2}{c}{\textbf{Sequence}} & \multirow{3}{*}{\shortstack{Time/\\subseq. (s)}} \\
\cmidrule(lr){4-15}\cmidrule(lr){16-18}\cmidrule(lr){19-20}
& & &
\multicolumn{7}{c}{\textbf{Motion}} &
\multicolumn{5}{c}{\textbf{Contact}} &
\multicolumn{1}{c}{\textbf{Motion}} &
\multicolumn{2}{c}{\textbf{Jerk}} &
\multicolumn{2}{c}{\textbf{Scene}} & \\
\cmidrule(lr){4-10}\cmidrule(lr){11-15}\cmidrule(lr){16-16}\cmidrule(lr){17-18}\cmidrule(lr){19-20}
& & &
MPJPE$\downarrow$ & H$_\mathrm{feet}\rightarrow$ & FS$\downarrow$ &
R@$3\uparrow$ & FID$\downarrow$ &
T$_\mathrm{goal}\downarrow$ & O$_\mathrm{goal}\downarrow$ &
C$_\mathrm{prec}\uparrow$ & C$_\mathrm{rec}\uparrow$ & C$_\mathrm{F1}\uparrow$ & C$_\%\rightarrow$ & P$_\mathrm{hand}\downarrow$ &
FID$\downarrow$ & PJ$\downarrow$ & AUJ$\downarrow$ &
Coll$_\%\downarrow$ & P$_\mathrm{max}\downarrow$ & \\
\midrule
GT & -- & -- & - & -3.30 & 0.29 & - & - & - & - & - & - & - & 0.49 & 1.7e-6 & - & 0.03 & 2.22 & 0.35 & 0.04 & -- \\
\midrule
w/o DNO & 75 & -- & 19.85 & -4.47 & 0.35 & 0.54 & 0.41 & 25.70 & 0.52 & 0.46 & 0.35 & 0.36 & 0.34 & 8.6e-5 & 0.81 & 0.13 & 2.15 & 0.45 & 0.05 & $\sim$1 \\
DNO (ours) & 75 & 150 & 19.16 & -4.28 & 0.33 & 0.54 & 0.45 & 24.41 & 0.48 & 0.49 & 0.36 & 0.38 & 0.35 & 8.2e-5 & 1.12 & 0.13 & 2.16 & 0.32 & 0.05 & 240.0 \\
DNO & 25 & 150 & 19.12 & -4.89 & 0.34 & 0.54 & 0.32 & 23.48 & 0.48 & 0.48 & 0.37 & 0.38 & 0.36 & 7.5e-5 & 0.53 & 0.14 & 2.12 & 0.32 & 0.05 & 85.3 \\
DNO & 25 & 100 & 19.10 & -4.52 & 0.33 & 0.52 & 0.28 & 24.64 & 0.47 & 0.47 & 0.37 & 0.38 & 0.35 & 6.7e-5 & 1.07 & 0.13 & 2.12 & 0.33 & 0.05 & 59.3 \\
DNO + skip & 25 & 100 & 18.87 & -5.13 & 0.35 & 0.53 & 0.39 & 23.45 & 0.43 & 0.47 & 0.35 & 0.37 & 0.35 & 8.2e-5 & 0.54 & 0.13 & 2.13 & 0.32 & 0.04 & 34.0 \\
\bottomrule
\end{tabular}
}
\end{table*}

\dy{\textbf{DNO configuration and runtime.}
Table~\ref{tab:supp_dno_runtime} reports the inference cost of DNO, averaged per subsequence on a single RTX A6000. Our default setting (75 DDIM and 150 optimization steps) takes 240\,s. Reducing the DDIM steps to 25 and the optimization steps to 100 lowers this to 85.3\,s and 59.3\,s, respectively, and further skipping DNO for subsequences that are already collision-free yields 34.0\,s, a 7$\times$ speedup. These reductions largely preserve quality: the fastest variant achieves MPJPE 18.87 and T$_\mathrm{goal}$ 23.45 (vs.\ 19.16 and 24.41) with the same Coll$_\%$ of 0.32. Truncated backpropagation through only a subset of denoising steps~\cite{clark2024directly} could further improve efficiency.}

\dy{\subsection{Results on the HIMO Dataset}
\label{sec_supp:himo}}
\begin{table*}[t]
\caption{\textbf{Results on the HIMO dataset~\cite{lv2024himo}}}
\label{tab:supp_himo}
\centering
\setlength{\tabcolsep}{3pt}
\resizebox{\textwidth}{!}{
\begin{tabular}{lccccccccccccc}
\toprule
Method & MPJPE$\downarrow$ & H$_\mathrm{feet}\rightarrow$ & FS$\downarrow$ & R@$3\uparrow$ & FID$\downarrow$ & C$_\mathrm{prec}\uparrow$ & C$_\mathrm{rec}\uparrow$ & C$_\mathrm{F1}\uparrow$ & C$_\%\rightarrow$ & P$_\mathrm{hand}\downarrow$ & FID$_\mathrm{tr}\downarrow$ & PJ$\downarrow$ & AUJ$\downarrow$ \\
\midrule
GT & -- & 1.71 & 0.02 & -- & -- & -- & -- & -- & 0.65 & 6.5e-4 & -- & 0.02 & 2.25 \\
\midrule
MDM & 9.59 & 1.23 & 0.05 & 0.80 & 0.31 & 0.71 & 0.63 & 0.62 & 0.56 & 7.5e-4 & 0.46 & 0.35 & 2.55 \\
CHOIS & 9.53 & 1.27 & 0.05 & 0.77 & 0.36 & 0.71 & 0.60 & 0.60 & 0.54 & 6.7e-4 & 0.58 & 0.34 & 2.50 \\
InterDiff & 10.93 & 0.37 & 0.05 & 0.80 & 0.33 & 0.68 & 0.50 & 0.52 & 0.45 & 9.5e-3 & 0.48 & 0.18 & 2.28 \\
DecHOI & 10.80 & 0.70 & 0.07 & 0.75 & 0.57 & 0.70 & 0.56 & 0.58 & 0.52 & 6.4e-4 & 2.32 & 0.61 & 2.95 \\
FlowMDM & 8.95 & 1.43 & 0.04 & 0.76 & 0.60 & 0.73 & 0.63 & 0.62 & 0.55 & 9.5e-3 & 0.87 & 0.01 & 2.30 \\
\rcol \ours~(ours) & 9.05 & 0.56 & 0.05 & 0.80 & 0.26 & 0.74 & 0.68 & 0.67 & 0.61 & 6.1e-4 & 0.47 & 0.09 & 2.19 \\
\bottomrule
\end{tabular}
}
\end{table*}

\dy{We additionally evaluate all methods on the HIMO dataset~\cite{lv2024himo}, a large-scale dataset for full-body human-object interaction with multiple objects. All methods are trained on HIMO and evaluated on its held-out split. Since HIMO sequences consist of multiple interaction segments, we adapt the dataset into a long-term HOI composition benchmark by treating 120 frames as a single subsequence, as the official release does not provide timeline annotations. Sequences are processed at 30 FPS without temporal downsampling, yielding 389 continuous test sequences with an average duration of 9.8\,s (up to 22.8\,s). Since HIMO does not provide 3D scene geometry, we disable the scene embedding, do not apply DNO during inference, and omit navigation-related metrics (Coll$_\%$, P$_\mathrm{max}$, T$_\mathrm{goal}$, O$_\mathrm{goal}$). \dyn{For R@3 and FID, we use the official evaluator released with HIMO.}
As shown in Table~\ref{tab:supp_himo}, \ours~achieves the tied-best performance on R@3 and the best performance on FID, contact quality (precision, recall, and F1), and transition smoothness (AUJ), while FlowMDM yields the lowest MPJPE and PJ. We attribute this performance variation to the distinct characteristics of the two datasets. HIMO primarily focuses on short-term (9.8\,s on average), object-centric interactions without scene boundaries, a setting where full-sequence non-autoregressive models may be effective. We thus view HIMO as a complementary benchmark: it shows that \ours~generalizes to shorter, unconstrained interactions, while ParaHome serves to evaluate the core challenges addressed by our formulation: long-horizon composition, transitions between multiple objects, and scene-aware navigation.}

\dyn{\section{Perceptual Study}
\label{sec_supp:user_study}}
\dyn{To evaluate semantic correctness, visual realism, and physical plausibility beyond quantitative metrics, we conducted a perceptual study comparing \ours~with InterDiff and FlowMDM.}

\dyn{\subsection{Setup}
\label{sec_supp:user_study_setup}}
\dyn{\noindent\textbf{Participants.} 25 participants who were not involved in this work took part as unpaid volunteers. All participants voluntarily agreed to participate in the study.

\noindent\textbf{Stimuli.} We randomly sampled 20 test sequences (fixed seed, no manual selection), each containing 3--4 consecutive interactions. All methods are rendered identically, so that only the motion differs.

\noindent\textbf{Procedure.} Each trial presents two videos side by side (\ours~vs.\ InterDiff or FlowMDM) in random left/right order, together with the task list. Participants answer three forced-choice questions on (1) \textit{semantic correctness}, whether the motion follows the task list with the right objects and order; (2) \textit{visual realism}, whether it moves like a real human; and (3) \textit{physical plausibility}, whether it respects physics (\eg, hand--object penetration, foot sliding, or passing through furniture). After each choice, they rate their confidence as \textit{clear}, \textit{subtle}, or \textit{almost a guess}. Each participant judged 10 pairs, five against each baseline, giving 750 judgments in total (125 per criterion and baseline). The full instructions and the trial screen are provided in Sec.~\ref{sec_supp:user_study_inst}.}

\dyn{\subsection{Results}
\label{sec_supp:user_study_results}}
\begin{table}[t]
\caption{\textbf{Perceptual study}}
\label{tab:supp_user_study}
\centering
\setlength{\tabcolsep}{3pt}
\small
\begin{tabular}{llcccc}
\toprule
Criterion & Baseline & All & Clear & Subtle & Almost a guess \\
\midrule
Semantic correctness & InterDiff & 96.0\% & 96.9\% (65) & 97.6\% (41) & 89.5\% (19) \\
 & FlowMDM & 70.4\% & 77.2\% (57) & 68.3\% (41) & 59.3\% (27) \\
\midrule
Visual realism & InterDiff & 95.2\% & 97.7\% (88) & 89.7\% (29) & 87.5\% (8) \\
 & FlowMDM & 80.0\% & 95.5\% (66) & 67.3\% (49) & 40.0\% (10) \\
\midrule
Physical plausibility & InterDiff & 96.0\% & 100.0\% (78) & 90.0\% (30) & 88.2\% (17) \\
 & FlowMDM & 78.4\% & 94.1\% (68) & 65.8\% (38) & 47.4\% (19) \\
\midrule
Average & InterDiff & 95.7\% & 98.3\% (231) & 93.0\% (100) & 88.6\% (44) \\
 & FlowMDM & 76.3\% & 89.5\% (191) & 67.2\% (128) & 51.8\% (56) \\
\bottomrule
\end{tabular}
\end{table}

\dyn{Table~\ref{tab:supp_user_study} reports the preference rate for \ours, overall and split by confidence (number of judgments in parentheses). \ours~is preferred over both baselines across all three criteria, in 95.2--96.0\% of the comparisons against InterDiff and 70.4--80.0\% against FlowMDM. Against FlowMDM, the preference reaches 89.5\% for clear judgments and decreases for subtle (67.2\%) and near-guess (51.8\%) ones, indicating that the preference for \ours~is strongest when participants perceive a clear difference. Against InterDiff, \ours~is preferred at all confidence levels (98.3\% / 93.0\% / 88.6\%).}

\dyn{\subsection{Instructions to Participants}
\label{sec_supp:user_study_inst}}
\begin{figure}[t]
\centering
\includegraphics[width=0.7\linewidth]{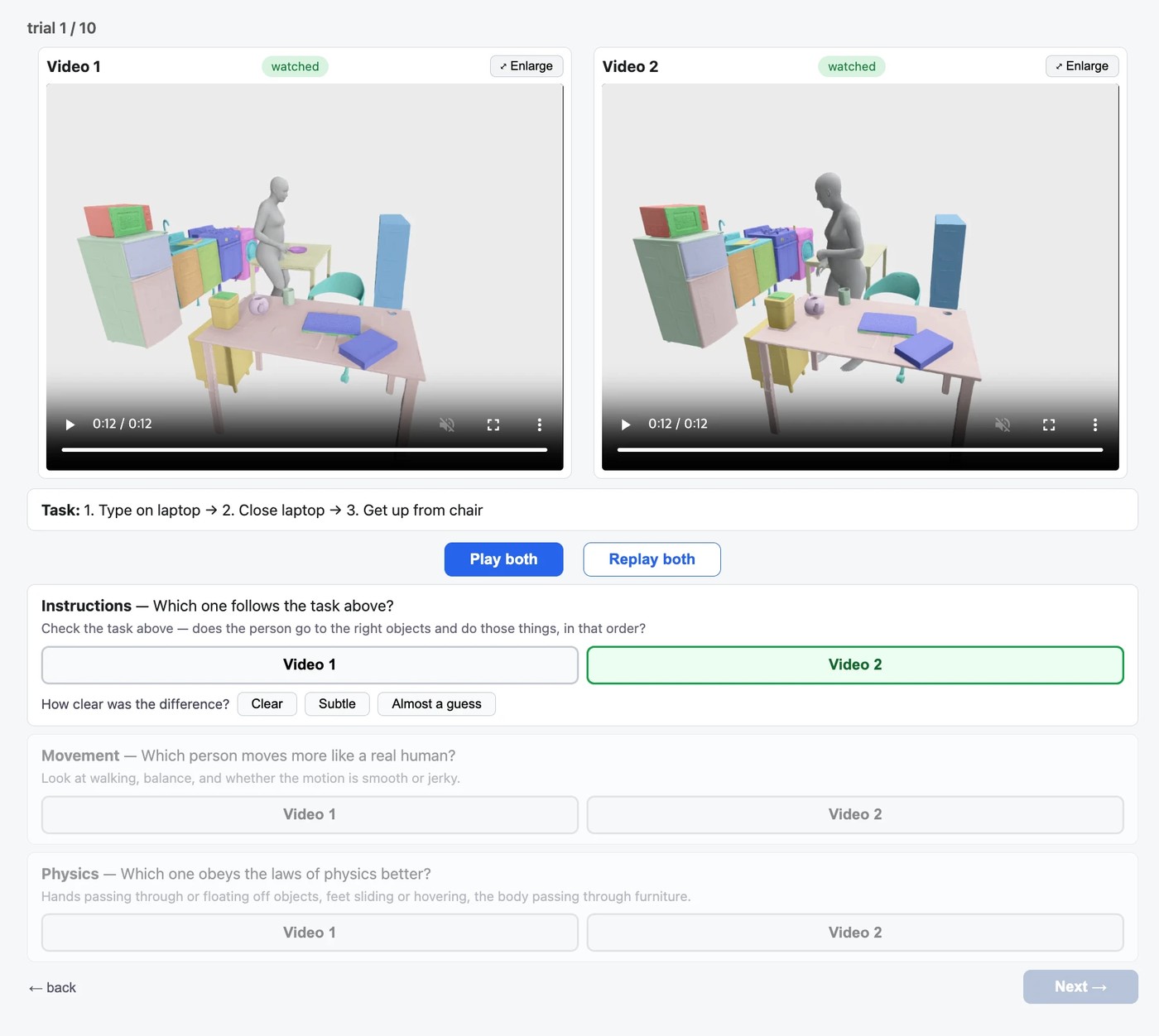}
\caption{\dyb{\textbf{Trial screen of the perceptual study}}}
\label{fig:supp_user_study_trial}
\end{figure}
\dyb{Before the study, participants were shown the following instructions.
\begin{quote}
\small
\textbf{Video comparison study.}\\
\textbf{What this is.} You'll watch short animations of a person doing everyday tasks in a room. Each screen shows two versions of the same task, made by two different systems. You pick which one is better in three ways. About 12 minutes, 10 screens.\\
\textbf{What to look for.}
\begin{itemize}
    \item Does the person do the listed actions, in order, with the right objects?
    \item Does the person move naturally, without jerky or sudden changes?
    \item Do hands hold objects, feet stay on the floor, and the body avoid furniture?
\end{itemize}
\textbf{What to ignore.} Colors, the plain grey body, blocky furniture, sharpness, lighting, camera angle. Both videos are rendered the same way---only the movement differs.\\
\textbf{Rules.} Watch both videos fully; the buttons unlock once they've played. You must pick one every time---there's no ``they're the same'' option.\\
\end{quote}}
\dyb{Each trial screen (Fig.~\ref{fig:supp_user_study_trial}) shows the two videos side by side together with the task list and asks the following three questions.
\begin{quote}
\small
\textbf{Instructions}---Which one follows the task above? Check the task above---does the person go to the right objects and do those things, in that order?\\
\textbf{Movement}---Which person moves more like a real human? Look at walking, balance, and whether the motion is smooth or jerky.\\
\textbf{Physics}---Which one obeys the laws of physics better? Hands passing through or floating off objects, feet sliding or hovering, the body passing through furniture.
\end{quote}
After each choice, participants are asked ``How clear was the difference?'' and select one of \textit{Clear}, \textit{Subtle}, or \textit{Almost a guess}.}

\section{Discussions}
\label{sec_supp:discussions}

\dy{\noindent\textbf{Limitations.}
\ours~takes object motions as input, treating them as a high-level spatio-temporal plan for human motion synthesis. In practice, such motions can be obtained from planning~\cite{yoshida2025generating, baik2026text} or object pose estimation~\cite{wen2024foundationpose, wen2023bundlesdf}, and \ours~remains effective under moderate object motion errors (Sec.~\ref{sec_supp:robustness}); jointly generating object and human motions from task-level goals is a promising future direction. In addition, the scene is represented as a 2D ground-plane map and the DNO objective acts on the pelvis trajectory, so fine-grained full-body collisions are not directly penalized. A per-joint objective reduces collisions further but degrades contact quality (Sec.~\ref{sec_supp:dno}), and a 3D voxel representation improves motion quality at about $4\times$ the training cost (Sec.~\ref{sec_supp:voxel}). Finally, DNO requires test-time optimization, which takes 240\,s per subsequence in our default setting and can be reduced to 34.0\,s with fewer steps (Sec.~\ref{sec_supp:dno}).}

\dy{\noindent\textbf{Failure cases.}
When the base model produces severe scene penetrations (\eg, passing through furniture), DNO may introduce minor floating or sliding artifacts while resolving them. These artifacts are not systematic: across the test set, foot-related metrics remain comparable after DNO, motion smoothness is preserved, and MPJPE improves (Table~\ref{tab:ablation}).}

\dy{\noindent\textbf{Broader impact.}
By generating realistic and temporally coherent long-term human--object interactions, \ours~can benefit embodied AI, AR/VR content creation, and digital human simulation. Since it does not explicitly model full physics or safety constraints, generated motions require additional verification before deployment on real-world robots.}

\section{Resources and License}
\label{sec_supp:resource}
\noindent\textbf{Compute resources.}
\dyn{We train and evaluate \ours~on a single NVIDIA RTX A6000 GPU; training for 200,000 iterations takes about 18.5 hours and requires about 29\,GB of GPU memory.} We implement \ours~in PyTorch.

\noindent\textbf{Licenses of existing assets.}
We use the ParaHome dataset~\cite{kim2025parahome}, which is released under the Creative Commons Attribution-NonCommercial-ShareAlike 4.0 International License (CC BY-NC-SA 4.0). For body and hand parameterization, we use the SMPL-X model~\cite{pavlakos2019expressive}, which is released under the Max Planck non-commercial scientific research license. \dyn{For the experiments in Sec.~\ref{sec_supp:himo}, we use the HIMO dataset~\cite{lv2024himo}, which is released under the HIMO license for non-commercial scientific research. We encode text with the pretrained CLIP model~\cite{radford2021learning}, which is released under the MIT License.} We also build upon publicly available code releases of MDM~\cite{tevethuman}, CHOIS~\cite{li2024controllable}, \dyn{DecHOI~\cite{jung2026decoupled},} InterDiff~\cite{xu2023interdiff}, and FlowMDM~\cite{barquero2024seamless} for our baselines, each used under the licenses provided in their respective repositories.

\end{document}